\pdfoutput=1
\documentclass{article}
\usepackage{microtype}
\usepackage{amsmath,amssymb,amsfonts}
\usepackage{graphicx}
\usepackage{textcomp}
\usepackage{xcolor}
\usepackage[hyphens]{url}
\usepackage{hyperref}

\usepackage[accepted]{mlsys2025}
\makeatletter
\let\mlsysmakecaption\@makecaption
\renewcommand{\printAffiliationsAndNotice}[1]{%
\stepcounter{@affiliationcounter}%
{\let\thefootnote\relax
\long\def\@makefntext##1{\noindent\raggedright##1}%
\footnotetext{#1\\
\forloop{@affilnum}{1}{\value{@affilnum} < \value{@affiliationcounter}}{%
\textsuperscript{\arabic{@affilnum}}\csname @affilname\the@affilnum\endcsname\\}%
Correspondence to: \mlsyscorrespondingauthor@text}}}
\makeatother

\usepackage[normalem]{ulem}
\usepackage{booktabs}
\usepackage{multirow}
\usepackage{makecell}
\usepackage{array}
\usepackage{subcaption}
\makeatletter
\AddToHook{begindocument/end}{\let\@makecaption\mlsysmakecaption}
\makeatother
\usepackage{tikz}
\usetikzlibrary{positioning,calc}

\usepackage{siunitx}
\usepackage{colortbl}
\usepackage{xspace}

\newcommand{\commentout}[1]{}

\definecolor{light-gray}{gray}{0.80}

\newcommand\appref{Appendix~\ref}

\newcommand\fref{Fig.~\ref}
\newcommand\tref{Table~\ref}
\newcommand\sref{\S~\ref}

\usepackage{amsthm}

\usepackage{amsthm}

\newtheorem{mytheorem}{Theorem}
\newtheorem{assumption}[mytheorem]{Assumption}
\newtheorem{lemma}[mytheorem]{Lemma}
\newtheorem{remk}[mytheorem]{Remark}

\newcommand{\minjia}[1]{{\color{red}\small [Minjia: #1]}}
\newcommand{\kartik}[1]{{\color{blue}\small [Kartik: #1]}}
\newcommand{\jiahuan}[1]{{\color{orange}\small [Jiahuan: #1]}}
\definecolor{darkgreen}{RGB}{0,100,0}
\newcommand{\kaidi}[1]{{\color{darkgreen}\small [Kaidi: #1]}}
\newcommand{\zihan}[1]{{\color{violet}\small [Zihan: #1]}}

\newcommand{\hide}[1]{}

\newcommand{\name}{FluidPD\xspace}
\newcommand{\instance}{FluidToken\xspace}
\newcommand{\cluster}{FluidRole\xspace}

\renewcommand{\emph}[1]{\textit{#1}}

\newcommand{\Require}{\REQUIRE}

\newcommand{\State}{\STATE}
\makeatletter
\newcommand{\For}[1]{\@ifnextchar\Comment{\mlsysforcomment{#1}}{\FOR{#1}}}
\def\mlsysforcomment#1\Comment#2{\FOR[#2]{#1}}
\makeatother
\newcommand{\EndFor}{\ENDFOR}
\newcommand{\If}{\IF}
\newcommand{\EndIf}{\ENDIF}
\newcommand{\Comment}[1]{\COMMENT{#1}}
\newcommand{\Return}{\textbf{return}\ }
\renewcommand{\algorithmiccomment}[1]{\hfill$\triangleright$ #1}

\let\cite\citep

\renewcommand{\minjia}[1]{}
\renewcommand{\kartik}[1]{}
\renewcommand{\jiahuan}[1]{}
\renewcommand{\kaidi}[1]{}
\renewcommand{\zihan}[1]{}

\mlsystitlerunning{FluidPD: In-Place Elasticity for P/D-Disaggregated LLM Serving}
\hypersetup{pdfauthor={Kartik Ramesh, Kaidi Fu, Zihan Zheng, Yi Zhao, Jiahuan Yu, Fabio Oliveira, Carlos H. A. Costa, Minjia Zhang},pdfsubject={}}

\begin{document}

\twocolumn[
\mlsystitle{\name: In-Place Elasticity for SLO-Aware Prefill-Decode Disaggregated LLM Serving}

\mlsyssetsymbol{equal}{*}

\begin{mlsysauthorlist}
\mlsysauthor{Kartik Ramesh}{equal,uiuc}
\mlsysauthor{Kaidi Fu}{equal,uiuc}
\mlsysauthor{Zihan Zheng}{uiuc}
\mlsysauthor{Yi Zhao}{uiuc}
\mlsysauthor{Jiahuan Yu}{uiuc}
\mlsysauthor{Fabio Oliveira}{ibm}
\mlsysauthor{Carlos H. A. Costa}{ibm}
\mlsysauthor{Minjia Zhang}{uiuc}
\end{mlsysauthorlist}
\mlsysaffiliation{uiuc}{University of Illinois Urbana-Champaign}
\mlsysaffiliation{ibm}{IBM Research}
\mlsyscorrespondingauthor{Minjia Zhang}{\mbox{minjiaz@illinois.edu}}

\vskip 0.3in
\begin{abstract}

Prefill-decode (P/D) disaggregation serves the two phases of LLM inference on separate workers, typically with a fixed prefill/decode worker ratio. Real workloads exhibit both short bursts and sustained shifts in the prefill-to-decode demand ratio, so a fixed split can violate latency SLOs while capacity sits idle in the other pool. Autoscaling adds capacity, but it reacts slowly, requires spare GPUs, and does not address short-timescale phase imbalance.  
We present \name, a P/D-disaggregated serving system with SLO-aware in-place elasticity.  \instance absorbs transient imbalance by offloading a bounded portion of prefill computation to decode workers that have slack. \cluster absorbs sustained imbalance by switching running workers between the prefill and decode roles in place, without model reload or engine restart. Both  are guided by  indices that expose the pressure in each phase before it causes SLO violations.
On workloads derived from production Azure traces and ShareGPT conversations with sustained shifts in the prefill-to-decode demand ratio, \name improves overall SLO attainment by up to 47.9 percentage points over the best static P/D split for each load,  without  additional workers.

\end{abstract}

]

\printAffiliationsAndNotice{\mlsysEqualContribution}

\section{Introduction}
\label{sec:intro}

Large language models (LLMs) power interactive applications such as chat assistants~\cite{openai2026chatgpt}, coding agents~\cite{chatgpt-codex,claude-cod,github2026copilot}, and tool-using workflows~\cite{react,openclaw,hermes-agent}, which have strict latency demands on both the first response  and the continuous token stream. 
Prefill-decode (P/D) disaggregation~\cite{patel2024splitwise,zhong2024distserve} meets these demands by placing the two phases on separate worker pools, each provisioned for its own execution pattern and Service Level Objective (SLO): prefill processes the prompt and largely determines Time-To-First-Token (TTFT), and decode generates output tokens  and determines Time-Per-Output-Token (TPOT).

However, this separation  makes capacity hard to move across phases.
Worker counts per phase are chosen ahead of time, often by offline profiling~\cite{patel2024splitwise,zhong2024distserve}, while production workloads exhibit both short bursts and sustained shifts in request mix~\cite{stojkovic2025dynamollm}: a burst of prompts can  overload the prefill workers~\cite{wang2025burstgpt}, and a shift between prompt-heavy coding requests and longer conversations changes the prefill-to-decode demand ratio~\cite{zhong2024distserve}.
Routing balances requests only within a phase's pool~\cite{zhong2024distserve,qin2024mooncake}, so the system can violate latency SLOs   
while capacity sits idle in the other pool.
Autoscaling in NVIDIA Dynamo~\cite{nvidia-dynamo}, AIBrix~\cite{aibrix2025}, and llm-d~\cite{llmd2025} adds GPUs instead of moving capacity between phases, and it is slow: in our Dynamo experiment, a new worker becomes ready a median of 111\,s after the scale-out decision (\sref{subsec:autoscaling-limitations}), long after a burst has ended. We call the missing capability \emph{in-place elasticity}: adapting the prefill/decode capacity  within the running deployment before changing its serving world size.

We present \name, the first P/D-disaggregated serving system to combine SLO-aware token-level work stealing for transient pressure with bidirectional in-place role switching for sustained phase imbalance, using only the running workers. For short-lived prefill pressure, \instance offloads a bounded suffix of selected prefill requests to their paired decode workers, so decode-side TPOT slack reduces the prefill backlog without changing  roles. When  pressure persists, \cluster switches workers between prefill and decode in place, without dropping in-flight requests.
Both mechanisms are guided by two predictive  signals derived from  engine state.
The Prefill Pressure Index (PPI) predicts each queued request's TTFT relative to its SLO, and the Decode Pressure Index (DPI) captures how close a decode worker is to its TPOT and KV-cache limits, including requests still waiting for a KV reservation. Aggregated over a pool, the two indices express demand in workers, which tells \cluster when moving one worker helps one phase without starving the other.
Unlike observed TTFT and TPOT, they expose pressure before it affects request latency.
Our contributions are:

\begin{itemize}
    \item We  design \instance, which limits stolen work to the TPOT slack of each decode step, and \cluster, whose role switch reloads no weights, captures no CUDA graphs, and re-establishes no KV-transfer  connections; PPI and DPI drive both (\sref{sec:design}).
    \item We implement \name on SGLang and evaluate it on workloads derived from production Azure traces and ShareGPT across four LLMs, including a 32B model with tensor parallelism. Under sustained shifts in the prefill-to-decode demand ratio, \name improves overall SLO attainment by up to 47.9 percentage points over the best static P/D split for each    load.
\end{itemize}

\section{Related Work}
\label{sec:bg}
Prior work improves LLM serving  through  batching and scheduling~\cite{fang2021turbotransformers,agrawal2024sarathiserve,yu2022orca}, KV-cache management~\cite{kwon2023pagedattention,zheng2023sglang,ge2024fastgen,li2024snapkv}, attention and decoding kernels~\cite{dao2022flashattention,ye2025flashinfer,prabhu2025vattention,aminabadi2022deepspeed,fastertransformer,wang2021lightseq},  offloading~\cite{xu2024pie,sheng2023flexgen}, parallelism~\cite{li2023alpaserve,pope2023efficiently}, speculative decoding~\cite{miao2024specinfer,stern2018blockwise}, and SLO-aware request scheduling~\cite{hong2025sola,bari2025slai,tang2025scorpio,kamath2025podattention,chen2026muxwise}, mostly for co-located prefill and decode; \name  targets P/D disaggregation,   
  which runs the two phases on separate workers. Splitwise~\cite{patel2024splitwise}, TetriInfer~\cite{hu2024tetriinfer}, and DistServe~\cite{zhong2024distserve} show that it improves goodput and TTFT and TPOT SLO attainment, and later work optimizes provisioning, scheduling, and KV-cache management~\cite{qin2024mooncake,hu2024tetriinfer}. 
Industry studies  show that the preferred P:D partition depends on the workload and changes over time~\cite{mitra2025pragmatic,li2025hetscale}, whereas these systems optimize a given P/D configuration and do not continuously adapt the running split.

Several systems adapt phase capacity by changing worker roles or serving groups. Arrow~\cite{wu2025arrow} reallocates stateless instances between prefill and decode, DOPD~\cite{liao2025dopd} moves instance allocation toward a predicted target P/D ratio,  P/D-Serve~\cite{jin2024pdserve} organizes P/D groups dynamically at deployment scale, and ThunderServe~\cite{jiang2025thunderserve} reschedules phase assignments  in heterogeneous clouds. Llumnix~\cite{sun2024llumnix} live-migrates requests across instances to rebalance load, and its open-source release adds an adaptive P/D mode that changes instance roles at runtime~\cite{llumnix-repo}. These systems adapt capacity at worker or serving-group granularity, but do not exploit token-level work stealing for short-lived phase pressure.

Other systems share capacity across the P/D boundary without  reconfiguring whole workers. TaiChi~\cite{wang2025taichi} and WindServe~\cite{feng2025windserve} mix prefill and decode execution within an instance. TokenScale~\cite{lai2025tokenscale} pairs a fast path, in which an offline-sized set of convertible decode workers runs whole burst prefills, with token-velocity-based autoscaling that changes the number of instances. Libra~\cite{ruan2026libra}, Kairos~\cite{arun2026kairos}, and PPD~\cite{li2026ppd}  move  execution across phases at token, request, or conversational-turn granularity, and Cronus~\cite{liu2025cronus} splits each prefill at a token boundary between a lower-end and a higher-end GPU. \instance differs in the combination of four choices: it moves a suffix of a queued prefill (not a whole request as in Kairos and TokenScale); a predicted TTFT-deadline miss triggers the move (not balancing execution time as in Libra and Cronus); the suffix runs on the request's paired decode worker within dedicated pools (not a mixed instance as in TaiChi and WindServe); and a per-step budget from that worker's TPOT slack protects decode. \cluster complements it by switching roles in place when pressure persists.

SGLang itself now offers related primitives. Since v0.5.21, an operator or the router can switch an instance between the prefill and decode roles after draining it to idle, on the Mooncake and Mori backends~\cite{sglang-pr28403}, following an earlier proposal~\cite{sglang-pr9325}. Open drafts let a decode worker prefill the uncached suffix of a request whose prefix it caches~\cite{sglang-pr34773,sglang-rfc45416}, as in PPD~\cite{li2026ppd}. PD-Multiplexing splits the SMs of one non-disaggregated instance between the two phases~\cite{sglang-pr11592}. Unlike SGLang's switch, \cluster switches a worker that is still serving, over NIXL: a P$\rightarrow$D worker hands its queue back to the router, a D$\rightarrow$P worker prefills while its decode requests finish, and weights, CUDA graphs, and KV-transfer connections stay in place. \name also adds a PPI/DPI controller that picks which worker to switch and when, work stealing bounded by the TPOT slack of every decode step, and joint control of both mechanisms.

\section{Observations and Opportunities}
\label{sec:obs}

\begin{figure}[t]
    \centering
    \includegraphics[width=1.00\linewidth]{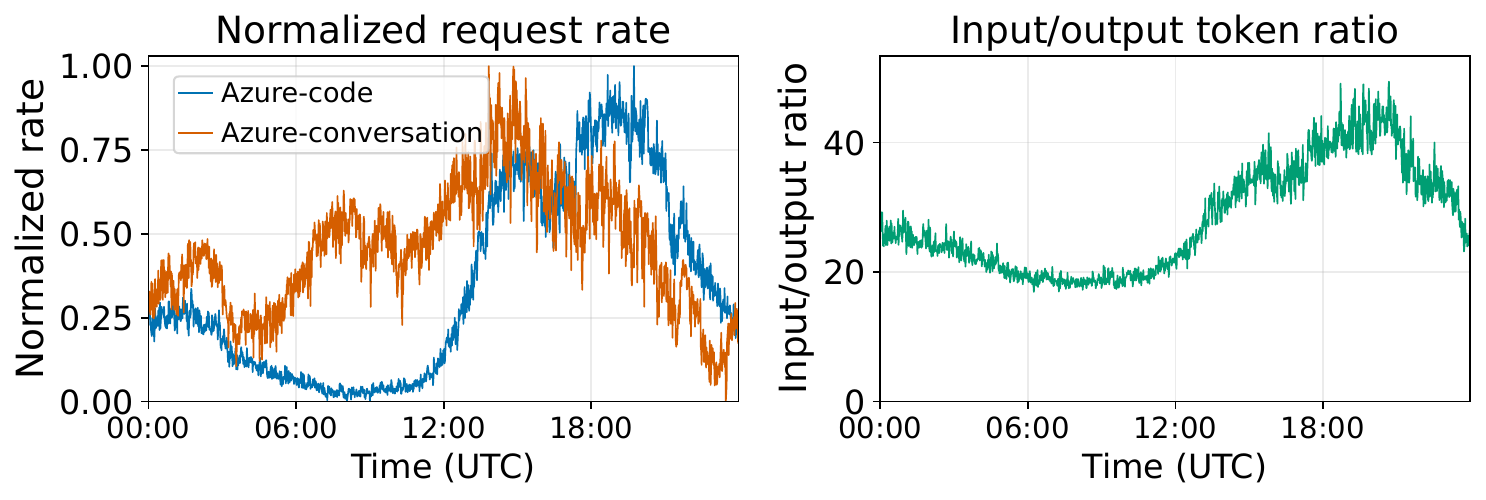}
    \caption{ Azure trace over one day (2024-05-14 UTC, 1-min bins).}
    \label{fig:pd-ratio}
\end{figure}

\subsection{Sustained Shifts Break Static P/D Splits}
\label{subsec:long-term-variation}
\begin{figure}[t]
    \centering
    \includegraphics[width=\columnwidth]{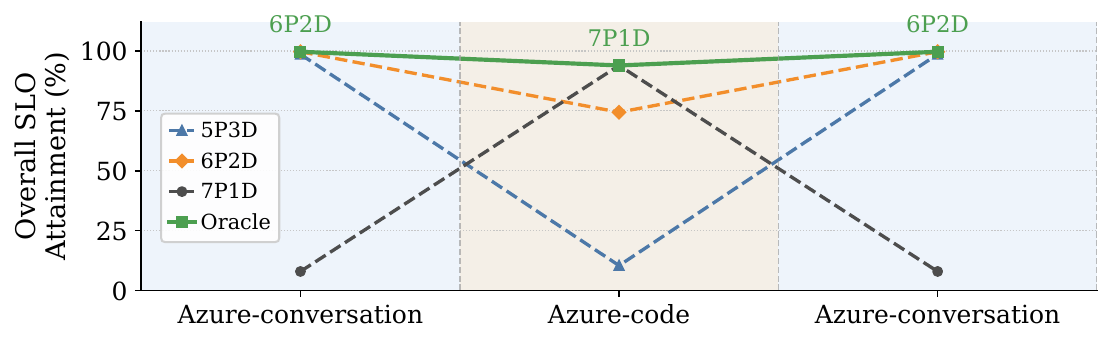}
    \caption{Sustained shifts favor different static P/D ratios (Llama-3.1-8B, 35\,RPS; stages run separately).}
    \label{fig:static-pd-adaptation}
\end{figure}

Prompt-heavy applications need relatively more prefill capacity and generation-heavy ones more decode capacity, so a change in the application mix shifts the P/D split that the workload needs.
\fref{fig:pd-ratio} shows one day of the Azure trace~\cite{stojkovic2025dynamollm}: coding requests, which are prefill-heavy, concentrate in business hours, while conversation requests, which carry relatively more decode work, arrive more evenly over the day. The aggregate input-to-output token ratio therefore varies from 16.9 to 49.4.  
A static  deployment binds worker capacity to phase roles, so it cannot follow this shift even when aggregate capacity suffices.
To measure the SLO impact, we take \emph{Azure (staged)}\ (\sref{subsec:eval-method}), three consecutive 10K-request windows in the order Azure-conversation, Azure-code,  Azure-conversation, and run each stage separately with Llama-3.1-8B at   35\,RPS under 5P3D, 6P2D, and 7P1D ($N$P$M$D denotes $N$ prefill and $M$ decode workers), with TTFT and TPOT SLOs of 2000\,ms and
100\,ms. No static split serves both workload types (\fref{fig:static-pd-adaptation}): 5P3D reaches 98.6\% SLO attainment on Azure-conversation but only 10.4\% on Azure-code, and 7P1D reaches 94.0\% on Azure-code but only 8.0\% on Azure-conversation; 6P2D, the best split for Azure-conversation (99.7\%), falls to 74.4\% on Azure-code. A per-stage oracle that picks the split with the highest overall SLO attainment for each window stays at or above 94.0\%, so the SLO loss comes from P/D imbalance rather than insufficient total capacity.

\subsection{Short Bursts Create Transient P/D Imbalance}
\label{subsec:short-term-variation}
\begin{figure}[t]
    \centering
    \includegraphics[width=\columnwidth]{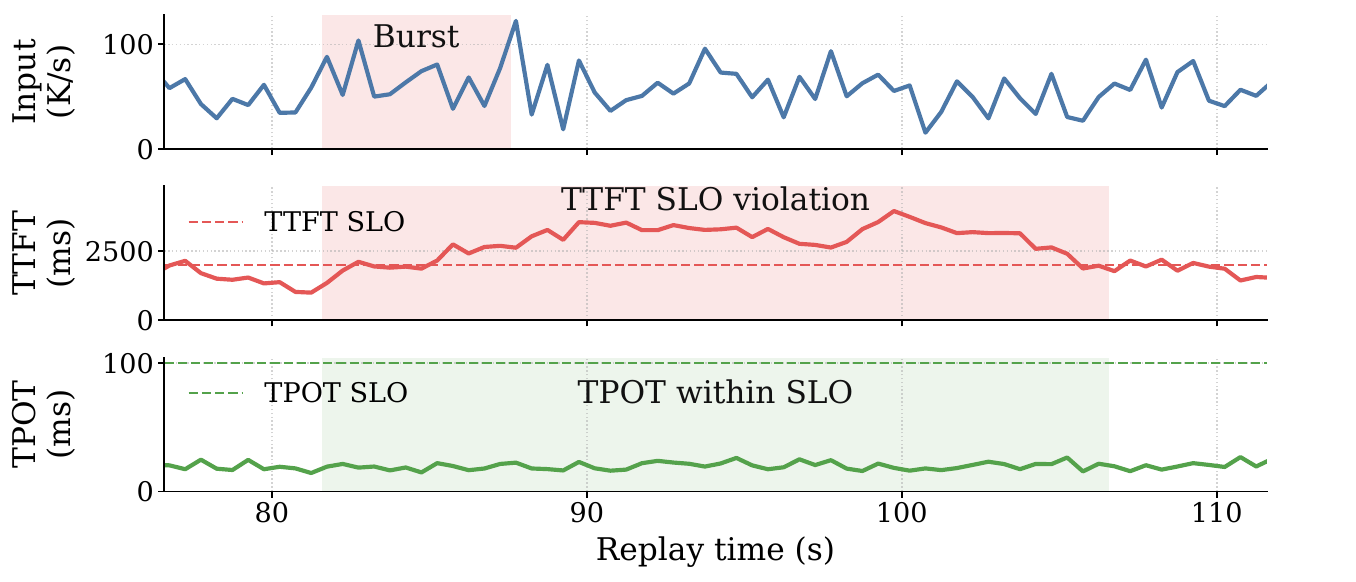}
    \caption{A burst violates TTFT  while TPOT stays within SLO.}
    \label{fig:short-term-burst-effect}
\end{figure}

Load also varies on a much shorter timescale. We call a brief interval in which arrival or token demand rises sharply a \emph{burst}; unlike a sustained shift, it need not change the long-term workload mix. 
We replay \emph{Azure (mixed)}, a 10K-request mix of Azure-code and Azure-conversation requests (\sref{subsec:eval-method}), at 30\,RPS with Llama-3.1-8B  under a static 4P4D deployment and the same SLOs (\fref{fig:short-term-burst-effect}).
During the marked burst, the input-token arrival rate rises from a median of 45K tokens/s over the preceding 5\,s to 66K tokens/s, with spikes above 100K tokens/s. TTFT exceeds its SLO as the prefill queue builds, and stays high after demand falls because new requests wait behind the backlog. TPOT stays within its SLO throughout: requests reach decode only after prefill and then generate one token per step, so the burst accumulates in the prefill queue while decode workers keep slack. A static split cannot lend this slack to prefill, and the burst may end before changing the  P/D split pays off. This calls for a fine-grained mechanism that borrows decode-side slack without changing worker roles.

 \phantomsection\label{subsec:autoscaling-limitations}Autoscaling is too slow for bursts and too costly for sustained shifts. In our Dynamo autoscaling runs (Llama-3.1-8B, SGLang backend, disaggregated planner, 8-GPU budget), an added worker becomes ready a median of 111\,s after the planner's scale-out decision, and never sooner than 100\,s; nearly all of  this time goes to initializing the  runtime, loading weights, and capturing CUDA  graphs. Even a scale-out decided when the burst begins adds capacity only after the burst and its 25\,s of TTFT violations in \fref{fig:short-term-burst-effect} have passed, and for a sustained imbalance, scaling out adds GPUs although the less-loaded pool already has slack (\appref{app:thirdparty}).

\section{System Design}
\label{sec:design}

\subsection{Overview}

\begin{figure}[t]
    \centering
    \includegraphics[width=\columnwidth]{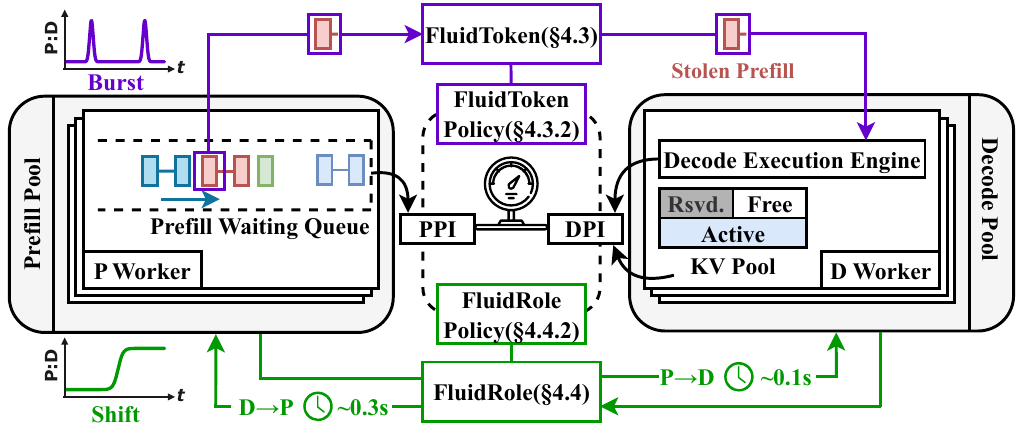}
    \caption{Overview of \name. \instance absorbs transient prefill bursts while \cluster handles sustained P/D demand shifts.}
    \label{fig:overall-arch}
\end{figure}

\fref{fig:overall-arch} shows the architecture of \name. On the standard P/D-disaggregated structure, in which a router assigns each request to a prefill/decode worker pair, \name adds two elasticity mechanisms.
\instance provides request-level adaptation (\sref{subsec:fluidtoken}): when a prefill worker predicts that queued requests will miss their TTFT SLO, it hands a suffix of selected requests ahead of them to their paired decode    workers, which run stolen suffix tokens only within the TPOT slack of their next decode step.  
\cluster provides worker-level adaptation (\sref{subsec:fluidrole}): when phase pressure persists, it switches workers between the prefill and decode roles in  place, keeping model weights, CUDA graphs, and KV-transfer
connections. Both are driven by two pressure signals (\sref{subsec:algorithm}), the Prefill Pressure Index (PPI) and the Decode Pressure Index (DPI). \instance acts in every prefill scheduling round and relieves transient  pressure first; \cluster acts only when the pool-level demand derived from the same indices stays outside the current P/D split for a confirmation   window. No prefill worker steals toward a decode worker that is switching roles, and a worker that still finishes residual decode requests after a switch is not switched back. Autoscaling remains for workloads that exceed the cluster's capacity.

\subsection{Prefill-Decode Pressure Signals}
\label{subsec:algorithm}

\name uses two phase-specific signals: the Prefill Pressure Index (PPI) predicts whether queued requests will miss their TTFT SLOs, and the Decode Pressure Index (DPI) captures how close a decode worker is to its TPOT and KV-cache limits. \instance uses both signals per request and per decode worker; \cluster aggregates them over each pool into demand measured in workers (\sref{subsec:fluidrole_policy}). Unlike observed TTFT and TPOT, which lag, both signals expose pressure before a violation, and they separate two causes of the same symptom: because the decode worker reserves KV before prefill starts~\cite{zheng2023sglang}, high TTFT can come from prefill overload or from a decode worker without KV headroom, even when prefill compute is available.

\noindent
\textbf{Prefill Pressure Index (PPI).}
A prefill worker packs its waiting queue,
in order, into batches of at most $B_{\mathrm{tok}}$ tokens (8192 in our runs), splitting a long prompt across batches
with chunked prefill.
An offline-calibrated linear model,
$\widehat{T}_{\mathrm{pre}}(\Lambda) = c_1 \Lambda + c_0$, predicts the duration of a prefill batch of
$\Lambda$
tokens. For request $i$ with arrival time $t_i^{\mathrm{arr}}$, prompt length $L_i$, and TTFT SLO $S_{\mathrm{TTFT}}$, let
$\widehat{T}^{\mathrm{done}}_i$ be the predicted time, from now, at which the batch holding $i$'s last
prompt token ends. Adding the time $i$ has already waited and the predicted KV transfer to its decode worker,
$\widehat{T}_{\mathrm{xfer}}(L_i) = x L_i$ for a fixed per-token transfer time $x$, gives its predicted TTFT, and $\mathrm{PPI}_i = \widehat{\mathrm{TTFT}}_i/S_{\mathrm{TTFT}}$ is its request-level pressure:
\begin{equation}
\widehat{\mathrm{TTFT}}_i = \left(t^{\mathrm{now}} - t_i^{\mathrm{arr}}\right)
  + \widehat{T}^{\mathrm{done}}_i + \widehat{T}_{\mathrm{xfer}}(L_i).
\label{eq:ppi-request}
\end{equation}
$\mathrm{PPI}_i \le 1$ exactly when $i$ is \emph{TTFT-schedulable}, i.e., its predicted prefill and
transfer complete before its deadline

$t_i^{\mathrm{ddl}} = t_i^{\mathrm{arr}} + S_{\mathrm{TTFT}}$. A request is \emph{lost} when it would miss its deadline even at the head of the queue.

At the worker level, $\mathrm{PPI}_p = W_p/S_{\mathrm{TTFT}}$ measures the prefill work $W_p$ a worker holds against the SLO:
\begin{equation}
W_p = R_p + \textstyle\sum_{k \in Q_p} c_1 L_k
+ c_0 \left\lceil \sum_{k \in Q_p} L_k / B_{\mathrm{tok}} \right\rceil ,
\label{eq:ppi-summary}
\end{equation}
where $R_p$ is the predicted remaining time of $p$'s launched batches and of a partially prefilled
request, and $Q_p$ is its waiting queue. Requests still waiting for a KV reservation on their decode
worker are not in $Q_p$; they count as decode demand (see below). $\mathrm{PPI}_p$ is the
predicted TTFT of a request that arrives now, as a multiple of the
SLO.

\noindent
\textbf{Decode Pressure Index (DPI).}
 
For each decode worker $d$, let $C_d^{\mathrm{KV}}$ denote its total KV-cache capacity, $C_d^{\mathrm{active}}$ the capacity held by running decode requests, and $C_d^{\mathrm{reserved}}$ the capacity reserved for admitted requests whose prefill or KV transfer has not finished.
Let $N_d^{\mathrm{active}}$ denote the current decode batch size, and $N_d^{\mathrm{TPOT}}$ the largest decode batch that satisfies the TPOT SLO $S_{\mathrm{TPOT}}$ under an
offline-profiled linear model of a decode step,
\[
\widehat{T}_{\mathrm{iter}}(N,C) = a_N N + a_C C + a_0 .
\]
A stolen suffix runs as its own forward pass, which we predict with the prefill model,
$\widehat{T}_{\mathrm{sfx}}(S) = \widehat{T}_{\mathrm{pre}}(S)$.
We decompose DPI into two components. The load-pressure component,
$\mathrm{DPI}_{d,\mathrm{load}} = \max(C_d^{\mathrm{active}}/C_d^{\mathrm{KV}},\ N_d^{\mathrm{active}}/N_d^{\mathrm{TPOT}})$,
captures the load imposed by running decode requests. The admission-pressure component captures KV
demand that has not been admitted yet. Requests for which $d$ cannot yet reserve KV wait at its door;
let $C_d^{\mathrm{pend}}$ be the KV they need (prompt plus maximum output). Then
$\mathrm{DPI}_{d,\mathrm{adm}} = (C_d^{\mathrm{active}} + w_d\,C_d^{\mathrm{pend}})/C_d^{\mathrm{KV}}$ with
$w_d = C_d^{\mathrm{active}}/(C_d^{\mathrm{active}} + C_d^{\mathrm{reserved}})$.
The weight $w_d$ counts waiting requests as decode demand only to the extent that running requests fill the pool: when prefill is the bottleneck, reservations fill it, and the queue at the door is a symptom of slow
prefill.
\subsection{\instance: Pressure-Guided Token Stealing}
\label{subsec:fluidtoken}

\subsubsection{Token-Stealing Execution}
\begin{figure}[t]
    \centering
    \includegraphics[width=\columnwidth]{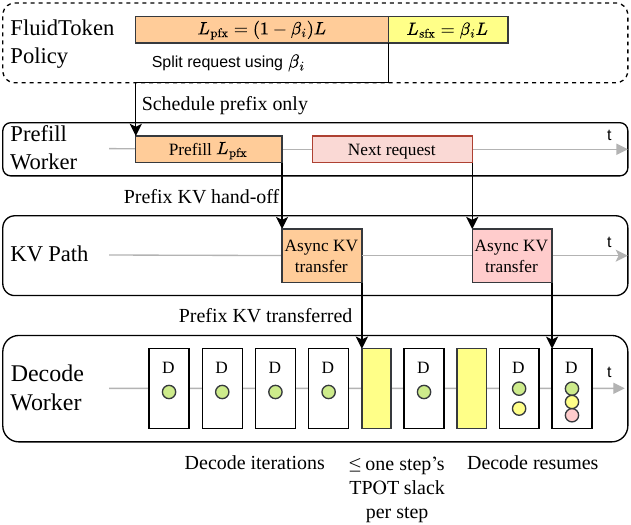}
    \caption{\instance execution flow.}
    \label{fig:fluidtoken-flow}
\end{figure}

\instance steals the suffix of a queued prefill onto the paired decode worker along the request's existing P/D flow (\fref{fig:fluidtoken-flow}), targeting the  bursts of \sref{subsec:short-term-variation}, where prefill queues build while decode-side slack remains.
For a request $i$ with prefill length $L_i$, \instance selects a \emph{stealing ratio} $\beta_i \in [0, 1)$ using the policy in \sref{subsec:fluidtoken_policy}, yielding a prefill-side prefix of $L_{\mathrm{pfx}}=(1-\beta_i)L_i$~tokens and a decode-side suffix of $L_{\mathrm{sfx}}=\beta_iL_i$~tokens.

After the prefill worker computes the prefix and transfers its KV states, the decode worker executes the suffix with the transferred KV as prior context, as in chunked prefill~\cite{agrawal2024sarathiserve}. Each suffix token sees the same causal mask and positions as in an unsplit prefill, so split execution computes the same KV states and outputs, up to floating-point differences from batch composition. After the suffix's last chunk (below), the decode worker emits the first token and the request joins the decode batch.

\begin{figure}[t]
    \centering
    \includegraphics[width=\columnwidth]{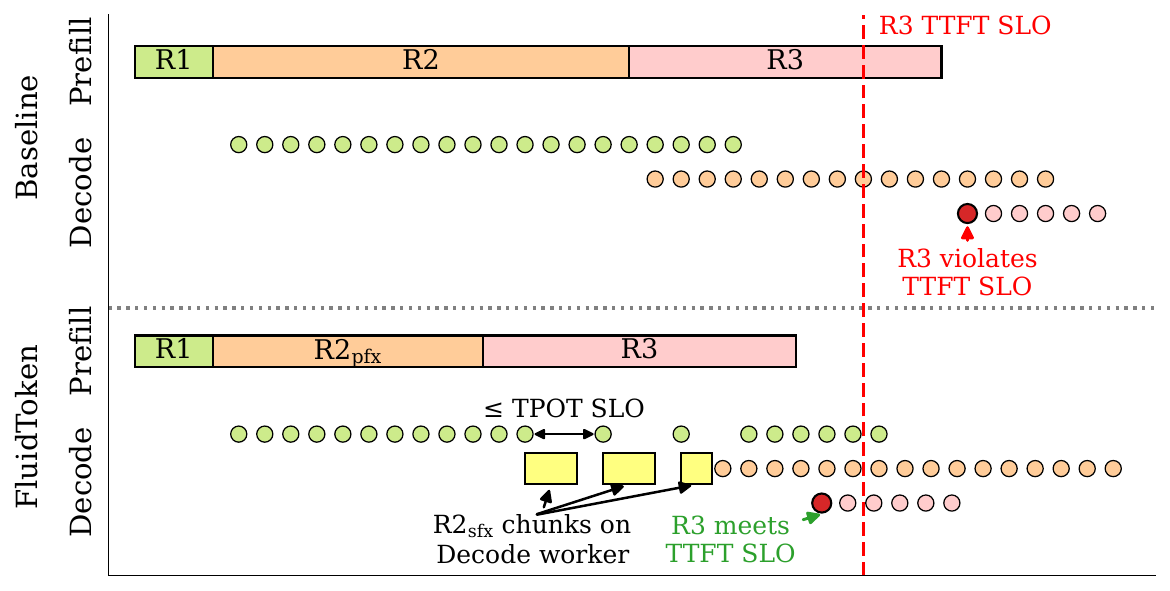}
    \caption{\instance splits  $R2$ so that $R3$ meets its TTFT SLO.}
    \label{fig:fluidtoken-example}
\end{figure}

As shown in \fref{fig:fluidtoken-flow}, \instance reuses SGLang's async P/D KV-transfer path~\cite{zheng2023sglang}, so both workers keep processing other requests while the prefix KV is transferred. \instance therefore redistributes rather than reduces prompt computation: it moves part of the queued prefill work to a decode worker with  slack, so the prefill worker begins subsequent requests earlier (\fref{fig:fluidtoken-example}).

Executing a stolen suffix has two  costs: its forward pass defers the next decode step of every running  request on the decode worker, and the suffix itself waits behind other suffixes there, which can make the stolen request late. It needs no extra KV, because the decode worker reserves KV for the whole prompt before prefill starts. \instance bounds the first cost per decode step and accounts for the second when it chooses which requests to split (\sref{subsec:fluidtoken_policy}).

\noindent\textbf{Decode-side execution.}\label{subsec:fluidtoken-decode}
A stolen suffix holds prompt tokens of the stolen request, which the decode worker runs as a prefill
forward over the transferred prefix KV. While the worker has running requests, it bounds the suffix
tokens it runs before each decode step by that step's TPOT slack: with the predictors of
\sref{subsec:algorithm}, it takes waiting suffixes first-in first-out, up to
\begin{equation}
S_{\mathrm{step}} = \max\left\{S : \widehat{T}_{\mathrm{sfx}}(S) \le
\eta\left(S_{\mathrm{TPOT}} - \widehat{T}_{\mathrm{iter}}(N, C)\right)\right\}
\label{eq:s-step}
\end{equation}
tokens ($\eta = 1$), where $N$ and $C$ are the size and context of its running batch. A suffix forward
delays the next decode step, so it counts toward the TPOT of the running requests, and toward the TTFT
of the stolen request. The bound holds per step, not on average, because of the inter-token latency (ITL),
the gap between two consecutive output tokens of a request: TPOT averages these gaps over the request, so
a whole stolen prefill run at once could keep TPOT within its SLO while every running request stalls for
the length of that prefill. A longer suffix is therefore split into chunks of at least $S_{\mathrm{chk}}$
tokens (256, \tref{tab:params}), each followed by a decode step; the minimum keeps suffixes
advancing when the decode step leaves no slack. Apart from this minimum, two consecutive decode steps are
at most one TPOT budget of predicted work apart, so the bound holds for every ITL, not only for TPOT. On
Azure (staged), wherever \name meets 90\% of the SLOs, its p99 ITL is at most 69\,ms on Llama-3.1-8B
and 117\,ms on Qwen3-14B, within their 100\,ms and 150\,ms TPOT SLOs (\sref{subsec:eval-main}).

\subsubsection{\instance Policy}
\label{subsec:fluidtoken_policy}

In every prefill scheduling round, \instance       chooses
stealing ratios $\{\beta_i\}$ that make late requests TTFT-schedulable without making others late and within the decode-side budgets.

\noindent\textbf{Stealing-aware TTFT constraint.}
A stolen suffix stays on its own request's TTFT critical path.  

For a request $j$ split into
$L_{\mathrm{pfx},j}$ and $L_{\mathrm{sfx},j}$ tokens,
$\widehat{T}^{\mathrm{done}}$ in \eqref{eq:ppi-request} is computed over prefix lengths. On the decode
worker $d$, $j$'s suffix runs after the $\Lambda_d$ suffix tokens queued ahead of it there: those already
waiting, those this prefill worker has in flight toward $d$, the shares the other prefill workers may
send (below), and suffixes split earlier in the same round. They drain at $S_{\mathrm{step}}(d)$ tokens
per decode step, so the suffix adds
$\widehat{T}^{\mathrm{dec}}_j = S_{\mathrm{TPOT}}(\Lambda_d + L_{\mathrm{sfx},j})/S_{\mathrm{step}}(d)$
to $j$'s TTFT ($0$ if $j$ is not split), and $j$ is TTFT-schedulable under stealing if
\begin{equation}
\begin{aligned}
\mathrm{PPI}^{\mathrm{steal}}_j = \big[ & \left(t^{\mathrm{now}} - t_j^{\mathrm{arr}}\right) + \widehat{T}^{\mathrm{done}}_j\\
 & + \widehat{T}_{\mathrm{xfer}}(L_{\mathrm{pfx},j}) + \widehat{T}^{\mathrm{dec}}_j \big] / S_{\mathrm{TTFT}} \le 1 .
\end{aligned}
\label{eq:stealing-schedulability}
\end{equation}

\noindent
\textbf{Decode-side stealing budget.}

Every $T_{\mathrm{rep}}$ (0.2\,s in our runs, \tref{tab:params}), each decode worker $d$ reports its running
batch ($N_d$, $C_d$) and the suffix tokens it has received but not executed to its prefill workers. From this report, prefill worker $p$ computes $S_{\mathrm{step}}(d)$ by \eqref{eq:s-step}, with the suffix cost corrected by $d$'s measured suffix
forwards, and lets stolen work toward $d$ stay outstanding for at most $k$ of its scheduling rounds:
\[
B_d = k\,S_{\mathrm{step}}(d)\,\widehat{T}_{\mathrm{round}}/S_{\mathrm{TPOT}}
\]
tokens, where $\widehat{T}_{\mathrm{round}}$ is the predicted duration of $p$'s next prefill batch. The budget
$B_{p,d}^{\mathrm{steal}}$ of $p$ is its share $B_d/|P_d|$ among the prefill workers $P_d$ currently
sending to $d$, minus its own outstanding suffix tokens toward $d$ (\appref{app:impl}). The budget is zero when $\mathrm{DPI}_{d,\mathrm{load}} \ge \mathrm{DPI}_{\mathrm{load}}^{\mathrm{high}}$, when $d$ is draining for a role switch, or when $d$'s report is stale. Because the decode worker
enforces $S_{\mathrm{step}}$ itself, $k$ only sets how much stolen work may queue at $d$; it never lets
stolen work exceed the TPOT slack of a decode step.

\noindent
\textbf{Request selection.}
Given its budgets, prefill worker $p$ chooses the stealing ratios $\beta_i \in [0,1)$ to minimize the number of requests predicted to miss their TTFT  SLO, where $d(i)$ is the decode worker paired with
 $i$:
\begin{equation}
\begin{aligned}
\min\nolimits_{\{\beta_i\}}\ & \textstyle\sum_{i} \mathbf{1}\!\left[\mathrm{PPI}^{\mathrm{steal}}_i > 1\right]\\
\text{s.t.}\ & \textstyle\sum_{i:\,d(i)=d} \beta_i L_i \le B_{p,d}^{\mathrm{steal}}\quad \forall d,\\
& \mathrm{PPI}^{\mathrm{steal}}_j \le 1 \ \text{if } \beta_j > 0 \text{ and } j \text{ is not lost}.
\end{aligned}
\label{eq:formation}
\end{equation}
The second constraint keeps every \emph{donor}, a request whose suffix is stolen, on time if it would otherwise meet its SLO.
Each prefill worker solves \eqref{eq:formation} for the requests at the head of its own
queue (up to $2S_{\mathrm{TTFT}}$ of predicted work
and at most 64 requests) with the greedy procedure in Algorithm~\ref{alg:fluidtoken-selection}.

\begin{algorithm}[t]
\small
\caption{\instance Stealing-Ratio Selection}
\label{alg:fluidtoken-selection} \setboolean{ALC@noend}{true}

\begin{algorithmic}[1]
\Require window $\langle R_1,\dots,R_n\rangle$ in queue order; budgets $B_{p,d}^{\mathrm{steal}}$
\State $\mathit{stolen}_j \gets 0\ \forall j$; \ $\mathit{budget}_d \gets B_{p,d}^{\mathrm{steal}}\ \forall d$
\For{each $R_i$ that is not lost, in queue order}
  \State \textbf{if} $\mathrm{PPI}^{\mathrm{steal}}_i \le 1$ \textbf{then continue}
  \State $X_i \gets$ token deficit of $R_i$; \ $\mathit{plan} \gets \emptyset$
    \For{$j \gets 1$ \textbf{to} $i-1$} \Comment{earliest donor first}
        \State $\mathit{take} \gets \min\{X_i,\ L_j{-}1{-}\mathit{stolen}_j,\ \mathit{budget}_{d(j)},\ \mathit{cap}_j\}$
    \State \textbf{if} $\mathit{take} < S_{\min}$ \textbf{and} $\mathit{stolen}_j = 0$ \textbf{then continue}
    \State $\mathit{stolen}_j \mathrel{+}= \mathit{take}$; $\mathit{budget}_{d(j)} \mathrel{-}= \mathit{take}$; $X_i \mathrel{-}= \mathit{take}$; add $(j,\mathit{take})$ to $\mathit{plan}$
    \State \textbf{if} $X_i \le 0$ \textbf{then break}
  \EndFor
  \State \textbf{if} $X_i > 0$ \textbf{then} undo $\mathit{plan}$ \Comment{$R_i$ cannot be saved}
\EndFor
\If{the queue holds a request that is not lost}
    \For{each lost $R_j$ in the window, in queue order}
    \State $\mathit{take} \gets \min\{L_j{-}1{-}\mathit{stolen}_j,\ \mathit{budget}_{d(j)},\ \mathit{cap}_j\}$
    \State \textbf{if} $\mathit{take} \ge S_{\min}$ \textbf{or} $\mathit{stolen}_j > 0$ \textbf{then} $\mathit{stolen}_j \mathrel{+}= \mathit{take}$; $\mathit{budget}_{d(j)} \mathrel{-}= \mathit{take}$
  \EndFor
\EndIf
\State \Return $\beta_j \gets \mathit{stolen}_j / L_j\ \forall j$
\end{algorithmic}

\end{algorithm}

Algorithm~\ref{alg:fluidtoken-selection} rests on a simple observation: under the linear prefill model, removing the same number of tokens from any request ahead of a late
request $i$ advances $i$ by the same amount, but removing them from an earlier request also advances
every request in between.
\instance therefore takes tokens from the earliest requests first, as long as
each donor stays on time. The \emph{token deficit} $X_i$ of $i$ is the smallest $X \ge 0$ with
$\widehat{T}^{\mathrm{done}}_i(X) + \widehat{T}_{\mathrm{xfer}}(L_i) \le t_i^{\mathrm{ddl}} - t^{\mathrm{now}}$,
where $\widehat{T}^{\mathrm{done}}_i(X)$ is the completion time of the batch holding $i$'s last prefix
token after $X$ tokens ahead of it
move to decode workers.

A request $j$ is \emph{eligible} to donate if it has unstolen prompt tokens, its decode worker has
budget left, and $\mathit{cap}_j$, the largest suffix that keeps $j$ (unless $j$ is lost) and the donors
behind $j$ on the same decode worker TTFT-schedulable by \eqref{eq:stealing-schedulability}, is
positive. No new split is smaller than $S_{\min} = \max(256, 2c_0/c_1)$ tokens, which keeps the
per-forward overhead of a suffix small relative to the work it moves. A late request takes tokens only
if its eligible donors can close its whole deficit, so no donor pays for a request that
stays late. Lost requests are never beneficiaries; after the late requests, they give up their own suffixes within
the remaining budget, since they have no deadline left to protect and every request behind them moves up. The
pass takes $O(n^2)$ time; under tensor parallelism, one all-reduce gives every rank the same split points (\appref{app:impl}).

\subsection{\cluster: In-Place Role Switching}
\label{subsec:cluster-level}
\label{subsec:fluidrole}

\subsubsection{Bidirectional Live-Role Migration}
\label{subsubsec:fluidrole-role-state}

\begin{figure}[t]
    \centering
    \makebox[\columnwidth][c]{\includegraphics[width=\columnwidth]{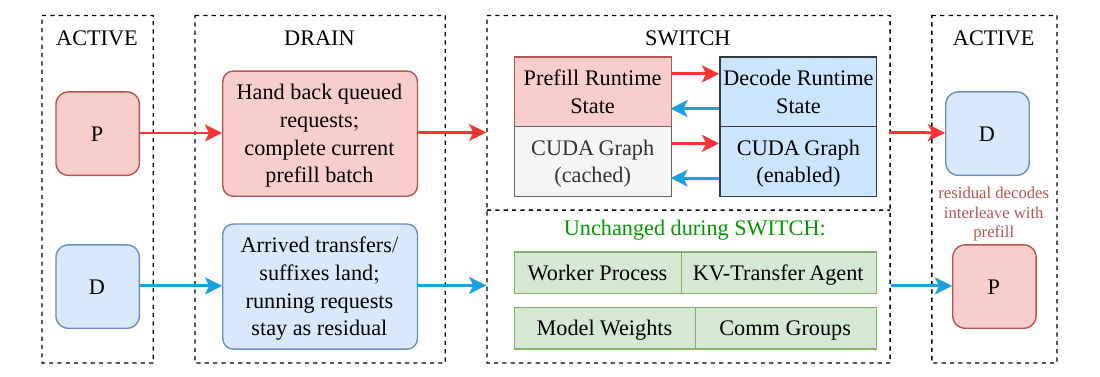}}
    \caption{\cluster role-switching state machine.}
    \label{fig:fluidrole-state}
\end{figure}

On a switch request, a worker moves through the state machine of \fref{fig:fluidrole-state}:
\textsc{Active}(\emph{old role}) $\rightarrow$ \textsc{Drain} $\rightarrow$ \textsc{Switch} $\rightarrow$ \textsc{Active}(\emph{new role}).
In \textsc{Drain},  it stops accepting new work and brings in-flight operations to  a safe boundary, which differs by direction (below). In \textsc{Switch},  \cluster activates the target role's queues, transfer handlers, and scheduler   
state.
\cluster keeps the worker's model placement and parallelism configuration, so a tensor-parallel worker reuses its  communication groups across the
switch. Every worker is launched role-neutral: with the same
configuration and KV page size, with decode CUDA graphs captured at startup, and with one KV-transfer agent that registers its KV pool once and connects to every peer before serving (\appref{app:impl}).
A switch therefore only replaces role state; it does not
restart the engine, reload weights, capture graphs, re-register memory, or reconnect to
peers.

\cluster switches a worker that is still serving without dropping requests: requests        that have not
started are reassigned to another worker of the same role, and requests in progress finish on the
selected worker, which holds up the switch differently in each direction.  
For a P$\rightarrow$D switch, the worker stops admitting  prefill requests,    hands its queued requests
back to the router, completes its current prefill batch, and flushes outstanding KV  transfers, so the switch takes about one prefill batch plus the remaining KV transfer.  A D$\rightarrow$P switch does not wait for active decode requests, which can keep a worker busy for
many seconds. The worker stops admitting decode requests and lets arrived KV transfers and stolen
suffixes finish. If its free KV then covers the maximum remaining output of its running decode requests
plus one prefill chunk, it switches to the prefill role at once and keeps them as \emph{residual}
requests; otherwise it lets them finish first. Until they finish, the worker is a prefill worker that
also decodes them: as in a co-located engine with chunked prefill, it interleaves their decode steps
with prefill batches, but it admits no new decode requests, sends the KV of each prefilled request to a
decode worker, and sizes each batch with \instance's predictors so that a decode step plus the batch
fits within $\eta_r S_{\mathrm{TPOT}}$ ($\eta_r=0.8$, 1.0 at TP=2; \tref{tab:params}). A D$\rightarrow$P
switch thus adds prefill capacity within about a second (0.29\,s at the median,
\sref{subsec:control-switching-overheads}), where draining would keep the worker out of service for
several seconds; its prefill share then grows as the residual requests finish, and it is not switched
back to decode while they run.

\subsubsection{\cluster Policy}
\label{subsec:fluidrole_policy}
\noindent
\cluster turns the indices into pool-level demand, measured in workers.
Prefill demand, $a_t = u^{-1}\sum_{p\in\mathcal{P}} (\lambda^{W}_p + \mathrm{PPI}_p)$, counts the prefill work arriving per second and the backlog that must be cleared within
one TTFT
SLO,\ 
where $\lambda^{W}_p$ is the predicted prefill time of the prompt tokens that arrived at $p$ per second
over the last 10\,s, i.e., the utilization the arriving work imposes on one worker, $\mathrm{PPI}_p$ is
from \eqref{eq:ppi-summary}, and $u = 0.9$ is the target utilization of one worker. Decode demand is
\[
\delta_t = u^{-1}\sum_{d\in\mathcal{D}} \max(\rho_d, \mathrm{DPI}_{d,\mathrm{adm}}),
\]
where $\rho_d$ is $d$'s load relative to the throughput it sustains at batch $N_d^{\mathrm{TPOT}}$: the larger of the
loads implied by its current batch and by its request
arrival rate. Unlike
$\mathrm{DPI}_{d,\mathrm{load}}$, which measures how full the batch is, $\rho_d$ measures throughput load
(\appref{app:impl}).  
With $n_P$ prefill and $n_D$ decode workers, \cluster switches a worker when one pool needs more
workers than it has and the other still meets its demand after giving one up:
\begin{equation}
\begin{aligned}
\text{D}{\rightarrow}\text{P}: &\quad a_t > n_P \ \wedge\ \min\left(\delta_t,\ \delta'_t\right) \le n_D - 1,\\
\text{P}{\rightarrow}\text{D}: &\quad \delta_t > n_D \ \wedge\ a_t \le n_P - 1.
\end{aligned}
\label{eq:fr-rules}
\end{equation} For D$\rightarrow$P, $\delta'_t$ is the decode demand projected onto
the $n_D-1$ workers that would
remain (\appref{app:a2}). $\delta_t$ is computed from each worker's current batch, which grows when stolen suffixes lengthen its
decode steps; the projection follows from arrivals alone. The two rules cannot hold at
the same time, and right after a switch the opposite rule cannot hold on the same signals, so the rules
leave a one-worker dead band and do not oscillate.
If the prefill pool needs more workers than it has
and the decode pool cannot spare one by either measure, neither rule holds and \cluster keeps the
split: the deployment lacks aggregate capacity, which admission control or autoscaling must address.
When the decode workers also reach their load limit, \instance stops as well, since a decode worker
at its load limit receives no stolen work.

\cluster evaluates the rules every 5\,s and switches when a rule holds twice in a row; the confirmation time and the dead band separate consecutive
switches, so there is no cooldown. It switches one worker at
a time: for D$\rightarrow$P the decode worker whose
longest remaining output is shortest, which bounds how long it keeps residual requests, and for P$\rightarrow$D the prefill worker with the least queued work and no residual
requests. It keeps at least one worker in each role and does not propose a worker that refused a switch again for 120\,s (\tref{tab:params}, \appref{app:params}).

\section{Evaluation}
\label{sec:eval}

\label{subsec:eval-method}

\noindent
\textbf{Implementation and testbed:} We implement \name on SGLang v0.5.4~\cite{zheng2023sglang}, extending its disaggregated runtime while keeping its P/D execution flow (decode-side KV-cache preallocation, asynchronous KV transfer over NIXL, CUDA graphs). \instance runs inside the prefill and decode schedulers; \cluster runs as a controller beside the router and issues role switches through it. Both mechanisms support tensor parallelism and any KV page size. The latency predictors are linear models fitted offline per model and tensor-parallel size, and they track measured prefill batches, stolen-suffix forwards, and decode steps with relative error within 20\% for most points (\appref{subsec:model-accuracy}). All experiments run on servers with 8$\times$ NVIDIA A100-80GB GPUs connected by NVLink, with one GPU per worker (TP=1) except for Qwen3-32B (TP=2, four workers): most on one AWS server, the rest on a second server of the same type (\appref{app:extra-main}). Every configuration uses 64-token KV pages, as in SGLang's P/D deployment examples~\cite{sglang-pd-lws}; with SGLang's default one-token pages, KV transfers slow down as the decode pool fills, and \appref{app:pagesize} repeats the Llama-3.1-8B configurations on the Azure workloads with them.

\begin{figure*}[!t]\centering
\begin{subfigure}{\textwidth}\centering
\includegraphics[width=\linewidth]{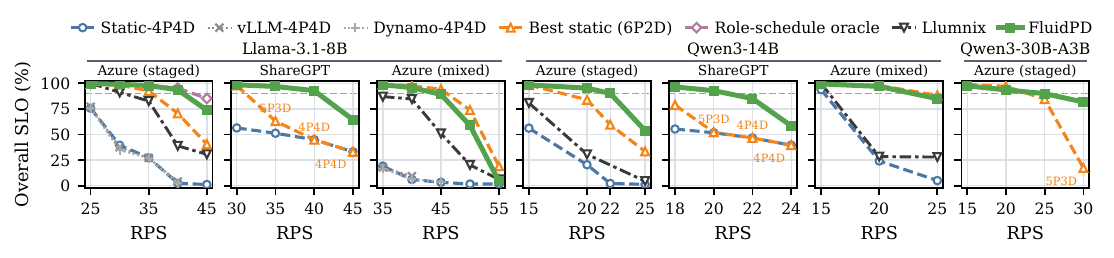}
\caption{Overall SLO attainment.}
\label{fig:eval-main-staged-azure-rps-sweep}
\end{subfigure}
\begin{subfigure}[t]{0.49\textwidth}\centering
\includegraphics[width=\linewidth]{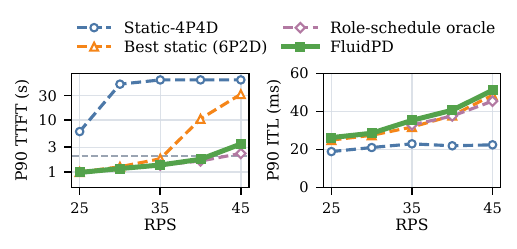}
\caption{90th-percentile latency, Llama-3.1-8B on Azure (staged).}
\label{fig:eval-main-p90}
\end{subfigure}\hfill
\begin{subfigure}[t]{0.49\textwidth}\centering
\includegraphics[width=\linewidth]{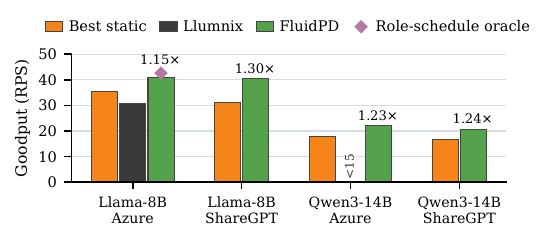}
\caption{Goodput.}
\label{fig:eval-goodput}
\end{subfigure}
\caption{Main results. (a) Overall SLO attainment on Azure (staged), ShareGPT (staged) and Azure
(mixed); Qwen3-14B with the 3000\,ms TTFT SLO and the 150\,ms TPOT SLO (100\,ms on Azure (mixed));
Qwen3-30B-A3B on Azure (staged) with the 1000\,ms TTFT SLO.
\name and Llumnix (\appref{app:llumnix}) start from 4P4D; the best static split is 6P2D except at the
labeled points (5P3D at every load for Qwen3-30B-A3B); the role-schedule oracle runs Llama-3.1-8B on Azure (staged) at 35--45\,RPS. vLLM-4P4D
and Dynamo-4P4D are static 4P4D deployments from our earlier experiments (\appref{app:thirdparty}).
(b) 90th-percentile TTFT, with a request that gets no first token counted as 60\,s, and inter-token
latency. (c) Goodput, the highest load at which at least 90\% of requests meet both SLOs, interpolated
between the loads of (a); labels give \name's goodput relative to the best static split.}
\label{fig:eval-main}
\end{figure*}

\noindent
\textbf{Models and workloads:} We evaluate \name on four   LLMs: Llama-3.1-8B, Qwen3-14B, Qwen3-30B-A3B (MoE), and Qwen3-32B (TP=2). The workloads derive from the week-long Azure-code and Azure-conversation traces of the Azure LLM Inference  trace~\cite{stojkovic2025dynamollm}. \emph{Azure (staged)} arranges 30K requests  as three 10K-request stages, Azure-conversation, Azure-code, and   
Azure-conversation; code requests have longer prompts and shorter outputs, so the stages shift the preferred P:D ratio. \emph{Azure (mixed)} contains 10K naturally interleaved requests from a bursty trace   
window (\sref{subsec:short-term-variation}). \emph{ShareGPT (staged)} replaces the two conversation stages of Azure (staged) with ShareGPT conversations~\cite{sharegpt} (\appref{app:extra-main}). Each request is the last user turn of a conversation, with the conversation up to that turn as its prompt, so these stages are decode-heavy: 1,085 prompt and 337 output tokens per request on average, against 1,631 and 108 for Azure-conversation. For Qwen3-14B and Qwen3-30B-A3B, each stage of the staged workloads holds 6K requests and Azure (mixed) 5K.
We replay these workloads with Poisson arrivals at controlled request rates, and every request generates exactly the output length its trace gives.

\noindent
\textbf{Baselines and configurations:} Our main baseline, \emph{Static}, runs the same SGLang build with both mechanisms disabled, so it differs from \name only in its fixed roles. We report Static at 4P4D, the split \name starts from, and at the best static split, the best of the splits we run at each load (\tref{tab:main}); it is 6P2D in every cell except Llama-3.1-8B on ShareGPT (staged) from 35\,RPS on, where it is 5P3D or 4P4D. For Llama-3.1-8B on Azure (staged), a role-schedule oracle knows the stage boundaries in advance and runs 6P2D in the conversation stages and 7P1D in the code stage, with \instance enabled. All SGLang configurations use identical worker settings, one load-aware router (\appref{app:impl}), and no prefix caching (hit rate below 1\%); the client abandons a request that has no first token after 60\,s. We also run Llumnix~\cite{sun2024llumnix,llumnix-repo} from 4P4D in its official adaptive-P/D deployment on the Azure workloads with the same client and SLOs (\appref{app:llumnix}), and show vLLM~0.25.1 and Dynamo~v1.3.0 as static 4P4D deployments from our earlier experiments, rescored the same way (\appref{app:thirdparty}). We   do not compare directly with TaiChi~\cite{wang2025taichi}, TokenScale~\cite{lai2025tokenscale}, Libra~\cite{ruan2026libra} (earlier DynaServe~\cite{ruan2025dynaserve}), or Kairos~\cite{arun2026kairos},  which had no public implementation at the time of evaluation.

\noindent
\textbf{Metrics:} Our primary metric, following~\cite{zhong2024distserve,patel2024splitwise,wu2025arrow}, is overall SLO attainment: the percentage of all issued requests that complete and meet both the TTFT and the TPOT SLO, where a request's TPOT is $(\mathrm{E2E}-\mathrm{TTFT})/(\text{output tokens}-1)$; failed and abandoned requests count as
violations. We also report the 90th-percentile TTFT and the 90th- and 99th-percentile ITL. In the main results, the TPOT SLO is 100\,ms for all models except Qwen3-14B (150\,ms on the staged workloads), and the TTFT SLO is 1000\,ms for Qwen3-30B-A3B, 2000\,ms for Llama-3.1-8B, and 3000\,ms for Qwen3-14B and  Qwen3-32B, following prior work~\cite{zhong2024distserve,wu2025arrow}; \appref{app:slo} varies the TPOT SLO, which sets \instance's stealing budget. Each configuration runs once (seed 1); seven cells near the knees, repeated with seed 2, differ by at most 3.4 points except one Qwen3-14B cell (\tref{tab:seeds}).

\subsection{Main Results}
\label{subsec:eval-main}

\fref{fig:eval-main-staged-azure-rps-sweep} compares \name with static 8-GPU deployments as the load grows. On Azure (staged), \name improves on the best static split at every load, and the gain grows near the knee: 33.2 points at 45\,RPS on Llama-3.1-8B (73.7\% versus 40.5\%) and 30.6 points at 22\,RPS on Qwen3-14B. No fixed split fits both stage types (\fref{fig:static-pd-adaptation}); \name moves from 4P4D through 5P3D to 6P2D, to 7P1D in the code stage, and back to 6P2D in the last stage. It stays within 1.7 points of the role-schedule oracle up to 40\,RPS; at 45\,RPS it trails the oracle by 11.5 points, mostly because it leaves 7P1D 40\,s into the last stage, while the oracle switches at the stage boundary.

On ShareGPT (staged), no fixed split serves both kinds of stage on Llama-3.1-8B from 35\,RPS on: two decode workers fall behind in the conversation stages (Static-6P2D meets the SLOs for 5.5\% of the requests at 40\,RPS), and splits with three or four decode workers lose the code stage. \name keeps 4P4D or 5P3D through the first stage, moves to 7P1D within a minute of the code stage, and returns to 5P3D within 40\,s of the last stage; it improves on the best static split by 47.9 points at 40\,RPS (92.8\% versus 45.0\% for 4P4D). On Qwen3-14B, it improves on it by 40.6 points at 20\,RPS (92.8\% versus 52.2\% for 5P3D). Measured as goodput, \name sustains 1.15--1.30$\times$ the load of the best static split on the same eight GPUs (\fref{fig:eval-goodput}). On every workload, \name stays above Static-4P4D at every load, and vLLM-4P4D and Dynamo-4P4D stay within 4.7 points of Static-4P4D: the 4P4D split, not the engine, limits them. Llumnix also stays above Static-4P4D, but in every Azure panel its attainment drops below 90\% at a lower load than \name's: on Llama-3.1-8B on Azure (staged), it reaches 38.9\% at 40\,RPS, against 93.6\% for \name. Against Dynamo's autoscaling, which may grow the same 6-GPU 3P3D start to eight GPUs, \name on six GPUs reaches 99.5\% and 98.0\% on Azure (mixed) and (staged) at 20\,RPS, against 9.2\% and 4.9\% (\appref{app:thirdparty}).

   Overall attainment follows TTFT attainment for every system, because shifts and bursts both first raise the pressure at the prefill workers (\appref{app:ttft-tpot}). \name's TPOT attainment stays at or above 98.5\% up to 40\,RPS on Llama-3.1-8B. At the highest loads it is 2.3--3.2 points below that of the best static split, and its p99 ITL is 111--120\,ms against 63--75\,ms: most of its gaps above 100\,ms fall while the decode worker runs a stolen suffix (\sref{subsec:fluidtoken-decode}). Under a strict-TTFT (1000/150\,ms) and a strict-TPOT (3000/50\,ms) SLO, \name keeps its TPOT attainment at or above 98.8\% and its TTFT attainment within 1.8 points of the best static split (\appref{app:slo}).

On Azure (mixed) (\fref{fig:eval-main-staged-azure-rps-sweep}), one split, 6P2D, is best at every load, and \name moves from 4P4D to it within 31\,s of the first request on Llama-3.1-8B. \name stays within 1.9 points of the best static split up to 40\,RPS on Llama-3.1-8B and within 3.3 points on Qwen3-14B, and above Static-4P4D at every load. On Llama-3.1-8B it is 4.9 points below at 45\,RPS and about 15 points below at 50 and 55\,RPS, where the best static split itself drops to 73.9\% and 19.0\%: requests that arrive in the first 100\,s, while \name moves to 6P2D and works off the queue built before the switch, account for the gap, and later requests meet their SLOs as often as under the best static split.

On Qwen3-30B-A3B (\fref{fig:eval-main-staged-azure-rps-sweep}), \name follows the stages of Azure (staged): from 25\,RPS on, it adds prefill workers in the code stage, up to 6P2D, and returns to decode-heavier splits afterwards. It exceeds the best static split by 5.5 points at 25\,RPS, reaches 81.7\% at 30\,RPS against 17.3\% for Static-5P3D, and stays within 3.5 points of it at 15 and 20\,RPS. On Azure (mixed), it keeps its 4P4D start and stays within 2.9 points of the best static split (\tref{tab:app-q30}). With tensor parallelism, \name serves Qwen3-32B on four TP=2 workers at 8\,RPS with 82.9\% and 91.0\% attainment on Azure (staged) and (mixed), against 74.3\% and 90.4\% for Static-3P1D (\appref{app:models}). \appref{app:extra-main} adds other starting splits and higher loads, and \appref{app:prodreplay} a replay of the production trace.

\begin{figure}[t]
    \centering
    \includegraphics[width=\linewidth]{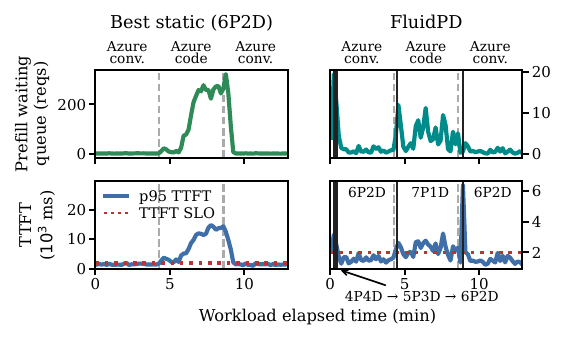}
    \caption{Llama-3.1-8B on Azure (staged) at 40\,RPS: the best
    static split (left) and \name from 4P4D (right), in 10\,s bins. Dashed lines mark stage boundaries, solid lines
    \name's role switches.}
    \label{fig:analysis-main-ttft-over-time}
\end{figure}

\noindent
\textbf{Why \name helps:}
We examine Llama-3.1-8B on Azure (staged) at 40\,RPS against the best static split, 6P2D (\fref{fig:analysis-main-ttft-over-time}). \name reaches 6P2D within 26\,s of the first request. When the code stage raises input-token demand, \cluster switches to 7P1D 15\,s into the stage; \name's prefill queue then stays at or below 12 requests, and 86.6\% of the code-stage requests meet both SLOs. The best static split cannot adjust: its prefill queue grows to about 280 requests, and only 26.2\% of the code-stage requests meet both SLOs. \name switches back to 6P2D 20\,s into the last stage.

\subsection{Component Ablation}\label{subsec:eval-ablation}
\begin{figure}[t]
    \centering
    \includegraphics[width=\linewidth]{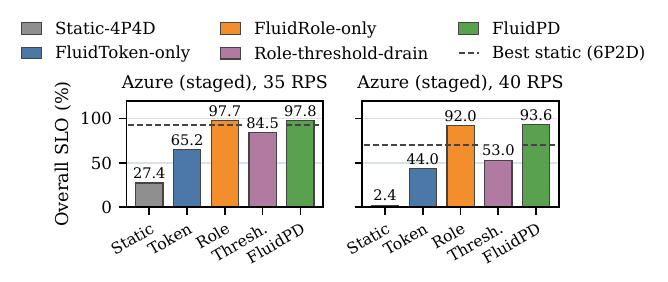}
    \caption{Component ablation on Llama-3.1-8B, starting from 4P4D, on Azure (staged). The dashed line marks the best static split (6P2D).}
    \label{fig:analysis-ablation}
\end{figure}

To isolate the contribution of each mechanism, we compare \name against Static-4P4D, \instance-only,  \cluster-only, and Role-threshold-drain on Llama-3.1-8B, all starting from 4P4D, on Azure (staged) at 35 and 40\,RPS (\appref{app:mixed} shows Azure (mixed)). Role-threshold-drain uses a threshold controller without stealing: it moves one worker, draining it, when the pressure indices stay beyond fixed thresholds for several consecutive intervals, and then waits out a cooldown.
\fref{fig:analysis-ablation} shows that \cluster provides the larger individual benefit: the workload needs more prefill workers than 4P4D provides for most of the trace, and only a role switch adds them, while \instance-only is bounded by the TPOT slack of the four decode  workers. Role-threshold-drain reaches 84.5\% and 53.0\%; at 40\,RPS it waits 45\,s before its first switch and never reaches 7P1D, whereas \cluster reaches 6P2D within 26\,s. \name performs best at both loads.  
Under pulses of 5 to 300\,s that add 18\,RPS of Azure-code requests to Azure (mixed) at 18\,RPS, \instance-only handles the short pulses and \cluster-only the long ones, while \name is the best of the three at every length; under a step that overloads both pools at once, \name reaches 49.8\%, against at most 35.5\% for either mechanism alone and 18.0\% for Static-4P4D (\appref{app:extra}).

\subsection{Control Parameters and Overheads}\label{subsec:control-switching-overheads}

\noindent
\textbf{Control parameters:}
\name uses the parameter values of \tref{tab:params} (\appref{app:params}) for all models, workloads, and loads. \instance gives stolen work the full TPOT slack of each decode step ($\eta=1$) and lets a stolen suffix stay outstanding at a decode worker for up to $k=3$ prefill rounds, the largest value in $\{2,3,4\}$ that kept every request's TPOT within its SLO and every decode-step gap below 0.2\,s on a Llama-3.1-8B burst workload. \cluster has a single target utilization, $u=0.9$, and a 10\,s confirmation time; because its rules compare demand with worker counts, they need no per-workload thresholds.

\noindent
\textbf{Overheads:} Computing PPI, DPI, and the stealing budgets uses scheduler metadata that SGLang already maintains plus the decode workers' reports. The  CPU-side overhead is 0.4\,ms per scheduling round for the stealing decision with Qwen3-32B at TP=2, plus 0.2\,ms to agree on the split points across ranks. SGLang's overlap scheduler pipelines CPU-side scheduling with GPU execution, so this overhead is largely hidden from the prefill and decode critical paths. \instance adds GPU work only when a stolen suffix executes on a decode worker, and this cost is included in all latency and SLO results.
\tref{tab:switch} (\appref{app:switchcost}) breaks down the cost of a role switch. An idle switch takes under 10\,ms, since it only replaces role state. Under load, a P$\rightarrow$D switch issued by \cluster takes 0.11\,s at the median and at most 0.99\,s, dominated by completing the current prefill batch and flushing outstanding KV transfers. A D$\rightarrow$P switch hands the worker back to the router as a prefill worker after 0.29\,s at the median and at most 1.5\,s with TP=1 (1.31\,s with TP=2); with a 100\,ms TPOT SLO, its residual decode requests keep a p99 ITL of 95\,ms and finish 17\,s later at the median. Draining the worker instead holds it for 2.2--11.5\,s.  \appref{app:correctness} exercises the switch path with forced races, residual requests, and output comparisons.

\section{Conclusion}
\name provides in-place elasticity for P/D-disaggregated serving: \instance steals bounded prefill work into decode-side TPOT slack, and \cluster switches worker roles in place when pressure persists. On workloads derived from Azure traces and ShareGPT with sustained shifts in phase demand, it improves overall SLO attainment by up to 47.9 percentage points over the best static P/D split for each load while largely preserving TPOT attainment except at the highest loads.

\section*{Acknowledgments}

This research was supported by the National Science Foundation (NSF) under Grant No. 2441601. The work utilized the Delta and DeltaAI system at the National Center for Supercomputing Applications (NCSA) and Jetstream2 at Indiana University through allocation CIS240055 from the Advanced Cyberinfrastructure Coordination Ecosystem: Services \& Support (ACCESS) program, which is supported by National Science Foundation grants \#2138259, \#2138286, \#2138307, \#2137603, and \#2138296. The Delta advanced computing resource is a collaborative effort between the University of Illinois Urbana-Champaign and NCSA, supported by the NSF (award OAC 2005572) and the State of Illinois. UIUC SSAIL Lab is supported by research funding and gift from IBM, Google, Amazon, and AMD.

\bibliographystyle{mlsys2025}
\bibliography{_s9.9_ref}

\section*{AI Usage Disclosure}
We used Claude Code (Anthropic) as an assistant in this work. The research questions, the design of \name, and the
evaluation plan were developed by the authors.
\begin{itemize}
\item \textbf{Code:} Claude Code helped write parts of the \name implementation on SGLang, the experiment harness,
and the scripts that analyze the results and draw the figures. The authors reviewed and tested all of this code.
\item \textbf{Experiments:} Claude Code orchestrated the experiment runs on the GPU servers: it queued the runs the
authors specified, monitored them, and collected their results. Every reported number comes from these measured
runs; no AI tool generated or edited measurement data.
\item \textbf{Writing:} Claude Code helped improve the grammar and clarity of the text and checked the bibliography
entries against their sources. The authors revised and verified all text and all cited sources.
\end{itemize}
The authors take full responsibility for the technical content and conclusions of this paper.

\clearpage
\newpage
\appendix
\raggedbottom %
\section{Implementation Details}
\label{app:impl}

\noindent\textbf{Role-neutral workers.} Every worker starts with the same engine arguments, memory
fraction, and chunked-prefill size, whatever its initial role. Decode CUDA graphs are captured at
startup on every worker, so the first P$\rightarrow$D switch does not capture graphs. The decode CUDA
graphs come out of the memory SGLang reserves for CUDA graphs and activations, so every worker has the
same KV-pool size.

\noindent\textbf{KV-transfer agent.} Each scheduler process (one per TP rank) owns one NIXL agent for
its lifetime, with one registration of its KV pool and of its auxiliary buffers, sized for the larger
of the two roles. At startup, every worker loads the transfer metadata of every other worker and is
loaded by them, and warms each connection with a 64-byte read per KV region, all before it reports
ready. A role switch replaces only the per-role state (queues and transfer tables). A peer that appears
later is loaded and warmed when it first registers, and a D$\rightarrow$P switch completes only after
the worker has loaded all of its decode peers.

\noindent\textbf{Handing back and residual requests.} On P$\rightarrow$D, queued prefill requests that
have not started, including those whose decode KV is already reserved, return to the router, which
re-dispatches them. After a D$\rightarrow$P switch, the residual decode requests keep their KV and their
rows in the request table, and the worker refuses a P$\rightarrow$D switch until they finish.

\noindent\textbf{Router.} The router polls every worker's load once per second and sends each request
to the less loaded of two randomly chosen workers in the required role. A worker whose last poll failed
or is overdue, or that has just switched and not yet answered a poll in its new role, is skipped.

\noindent\textbf{Metrics for \cluster.} Prefill workers publish $W_p$ and a counter of arrived prompt
tokens; decode workers publish counters of arrived requests, prompt tokens and maximum output tokens,
and the KV demand of the requests waiting at their door. Each worker publishes at most every 0.5\,s,
also when idle, with a timestamp; \cluster ignores samples older than 2\,s. The batch term and the
arrival term of $\rho_d$ agree in steady state by Little's law; the arrival term leads when the load has
just changed, or when a prefill bottleneck holds the decode batch below the arriving demand. $\rho_d$
measures throughput load because a decode step grows with the batch, so a half-full batch can already
carry most of a worker's throughput.

\noindent\textbf{Projected decode demand.}\phantomsection\label{app:a2} For a D$\rightarrow$P proposal with $m = n_D - 1 \ge 2$,
let $j$ be the decode worker \cluster would switch. Each remaining worker $i$ takes an equal share of
$j$'s request stream, $\lambda'_i = \lambda_i + \lambda_j/m$, and of $j$'s running batch, KV reservations
and waiting KV demand; its mean prompt and output lengths $\bar{L}'_i$ and $\bar{O}'_i$ are the
arrival-weighted means of $i$'s and $j$'s. With the decode-step model
$\widehat{T}_{\mathrm{iter}}(n) = a_0 + \kappa'_i n$, $\kappa'_i = a_N + a_C c'_i$, where $c'_i$ is the
merged context per running request, the output-token rate $X'_i = \lambda'_i(\bar{O}'_i - 1)$ gives the
steady-state batch
\begin{equation}
n'_i = \frac{X'_i a_0}{1 - X'_i \kappa'_i} \qquad (n'_i = \infty \text{ if } X'_i \kappa'_i \ge 1),
\end{equation}
the batch at which the worker completes output tokens as fast as they arrive. From $n'_i$ follow the
load $\rho'_i$ and the admission pressure $\mathrm{DPI}'_{i,\mathrm{adm}}$ in the same form as $\rho_d$ and
$\mathrm{DPI}_{d,\mathrm{adm}}$, and $\delta'_t = \frac{1}{u}\sum_i \max(\rho'_i, \mathrm{DPI}'_{i,\mathrm{adm}})$.
With fewer than two remaining decode workers, \cluster does not project and $\delta'_t = \infty$.

\noindent\textbf{\instance bookkeeping.} Each decode worker's report lists the suffix tokens it has
received but not executed, in total and per sender. $P_d$ holds $p$ and every prefill worker that has
requests waiting at $d$ or has sent $d$ a suffix in the last 5\,s. A prefill worker charges a stolen suffix against its share when
it commits the split, keeps charging it after the transfer completes until the decode worker's report
reflects it (judged by the decode worker's last transfer poll), and stops when the decode worker has
executed it. Stolen suffix work does not enter the decode-side metrics that \cluster reads.

\noindent\textbf{Decode-side execution and online suffix cost.} On a decode worker, requests whose KV has just arrived join the
running batch at the start of the next iteration that runs a forward. A decode worker times each
stolen-suffix forward with CUDA events, without synchronizing, skipping the first forward after startup
or a switch, and keeps an exponentially weighted average (weight 0.2) of measured over predicted time,
$r_d$, which its reports carry. Prefill workers use it after three samples and only to make the suffix
cost larger: they scale the suffix cost by $\max(1, r_d)$. The decode worker's own per-step cap uses the
predicted cost, so all TP ranks cut identical chunks.

\noindent\textbf{Tensor parallelism.} With TP$>$1, rank~0 of a prefill worker computes the split
points, since they depend on clocks and decode reports, and one all-reduce per scheduling round gives
all ranks the same split points; the decode side is deterministic given its queues.

\noindent\textbf{KV pages.} Both mechanisms run at any KV page size $P$. A split point is a whole
token: the prefill worker sends the prefix as $\lceil L_{\mathrm{pfx}}/P \rceil$ pages, the last one
partly filled, and the decode worker's first suffix chunk writes the rest of that page before any
attention reads it. Suffix chunks may end at any token, because the decode worker reserves the pages of
the whole prompt when it admits the request. During a D$\rightarrow$P transition, prefill admission
keeps back exactly the pages that the residual requests will still take, plus one page, and rounds
the size cap of each prefill batch down to whole pages.

\noindent\textbf{Control parameters.}\phantomsection\label{app:params} \tref{tab:params} lists the control parameters
of \name (\sref{subsec:control-switching-overheads}).

\begin{table}[t]
\centering
\caption{\name control parameters.}
\label{tab:params}
\vskip 0.1in
\footnotesize
\begin{tabular}{@{}l l >{\raggedright\arraybackslash}p{0.47\columnwidth}@{}}
\toprule
Parameter & Value & Role \\
\midrule
$\eta$ & 1 & share of a decode step's TPOT slack for stolen work, \eqref{eq:s-step} \\
$k$ & 3 & prefill rounds a stolen suffix may stay outstanding \\
$\mathrm{DPI}^{\mathrm{high}}_{\mathrm{load}}$ & 0.6 & no stealing toward a decode worker at or above it \\
Window & $2S_{\mathrm{TTFT}}$ & queue head examined by Algorithm~\ref{alg:fluidtoken-selection}, at most 64 requests \\
$S_{\min}$ & $\ge 256$ & smallest new split, $\max(256, 2c_0/c_1)$ tokens \\
$S_{\mathrm{chk}}$ & 256 & smallest suffix chunk a decode worker runs in one forward \\
$T_{\mathrm{rep}}$ & 0.2\,s & decode report period; stale after 0.6\,s, or 1\,s unchanged \\
$u$ & 0.9 & target utilization of one worker in $a_t$, $\delta_t$ \\
Rate window & 10\,s & arrival rates in $\lambda^{W}_p$ and $\rho_d$ \\
Evaluation & 5\,s $\times$ 2 & \cluster confirmation time of 10\,s \\
Projection & $n_D-1 \ge 2$ & D$\rightarrow$P may use the projected demand $\delta'_t$, \eqref{eq:fr-rules} \\
Sample age & $\le 2$\,s & older samples: no decision \\
Role minimum & $\ge 1$ & workers \cluster keeps in each role \\
$\eta_r$ & 0.8 & residual decode step + prefill batch $\le \eta_r S_{\mathrm{TPOT}} - 4$\,ms; 1.0 at TP=2 \\
Back-off & 120\,s & a worker that refused a switch is not proposed again \\
\bottomrule
\end{tabular}
\end{table}

\begin{figure}[!t]
    \centering
    \includegraphics[width=\linewidth]{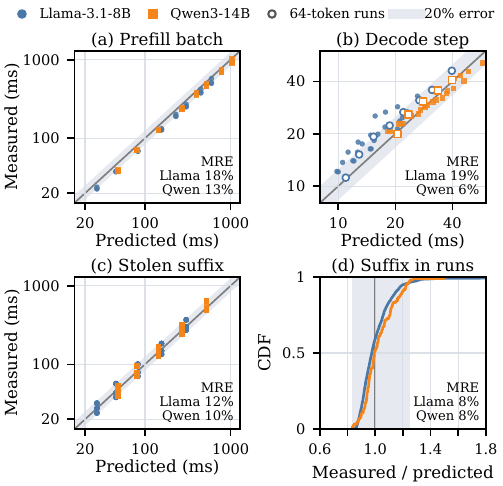}
    \caption{Accuracy of the latency models used by \name on our testbed. Filled markers: isolated forwards with one-token pages. Hollow markers in (b): median decode steps in the static runs with 64-token pages. (d): sampled stolen-suffix forwards in the runs with 64-token pages. MRE is the mean relative error; the band marks 20\% error.}
    \label{fig:analysis-model-accuracy}
\end{figure}

\noindent\textbf{Predictor accuracy.}\phantomsection\label{subsec:model-accuracy} \name uses lightweight latency predictors to estimate prefill batches, stolen-suffix forwards, and decode steps. They are linear models fitted offline per model and tensor-parallel size from single-worker profiling runs with one-token pages; for Llama-3.1-8B and Qwen3-14B, these runs used an A100-40GB GPU and mixed prefill-decode batches without CUDA graphs. On A100 GPUs, SGLang's attention kernels index the KV cache token by token at every page size, so the same predictors serve both page sizes. Prefill workers correct the suffix cost online with $r_d$ when the measured forwards are slower than predicted. \fref{fig:analysis-model-accuracy} compares predicted and measured execution times on our testbed. The predictions track the measurements with relative error within 20\% for most points, and 96\% of the stolen-suffix forwards sampled in our runs fall within 20\% of the prediction. This accuracy is sufficient for online control because \name uses the predictions to enforce conservative stealing and role-switching constraints, rather than to estimate exact end-to-end request latency.

\section{Additional Models}
\label{app:models}

\noindent This appendix gives the full results of Qwen3-30B-A3B, a mixture-of-experts model, and of
Qwen3-32B, which we serve with TP=2. Both use 64-token pages and the router, timeout and SLOs of
\sref{subsec:eval-method}. For Qwen3-32B, each stage of Azure (staged) holds the first 2K requests of the
corresponding 10K-request stage, and Azure (mixed) holds its first 5K requests.

\noindent\textbf{Qwen3-30B-A3B.} \tref{tab:app-q30} compares \name, which starts from 4P4D, with the
static 5P3D and 6P2D splits (\sref{subsec:eval-main}). After the code stage of Azure (staged), \name
returns to decode-heavier splits, down to 3P5D at 30\,RPS. In these runs, \name's decode-step predictor
uses the engine's profile, whose intercept, 45\,ms per step, is close to a decode step without CUDA graphs
(52\,ms measured) rather than with them (11\,ms). With decode costs measured on the A100 with CUDA graphs,
the stealing budget at the runs' median decode batches grows from at most 153 to 678--1,183 tokens per
step, and \name (column (m), second server) gains 2.0--2.3 points at 25 and 30\,RPS on Azure (staged)
and at 25\,RPS on Azure (mixed), and stays within 1.0 point elsewhere.

\begin{table}[t]
\centering
\caption{Overall SLO attainment (\%) of Qwen3-30B-A3B. \name starts from 4P4D.
Best is the better static split at that load, and $\Delta$ is \name minus Best in points, from unrounded values; a dash marks
a split we do not run at that load, and $^*$ the only static split we run at that load. (m): \name with
decode costs measured on the A100.}
\label{tab:app-q30}
\vskip 0.1in
\footnotesize
\setlength{\tabcolsep}{4pt}
\begin{tabular}{@{}l r r r r r l r@{}}
\toprule
Azure & RPS & \name & (m) & 5P3D & 6P2D & Best & $\Delta$ \\
\midrule
staged & 15 & 97.5 & 97.5 & 98.9 & 98.5 & 5P3D & $-1.4$ \\
 & 20 & 93.6 & 93.8 & 97.0 & 27.9 & 5P3D & $-3.5$ \\
 & 25 & 89.9 & 92.1 & 84.5 & 15.6 & 5P3D & $+5.5$ \\
 & 30 & 81.7 & 84.0 & 17.3 & -- & 5P3D$^*$ & $+64.4$ \\
\midrule
mixed & 10 & 99.7 & 99.7 & -- & 99.8 & 6P2D$^*$ & $-0.1$ \\
 & 15 & 98.9 & 97.9 & 99.2 & 99.4 & 6P2D & $-0.5$ \\
 & 20 & 96.6 & 96.1 & 98.5 & 22.5 & 5P3D & $-2.0$ \\
 & 25 & 91.0 & 93.0 & 93.9 & 7.4 & 5P3D & $-2.9$ \\
\bottomrule
\end{tabular}
\end{table}

\noindent\textbf{Qwen3-32B with TP=2.} With two GPUs per worker, the node holds four workers. \name and
\cluster-only start from 2P2D, and \tref{tab:app-q32} compares them at 8\,RPS with Static-3P1D, the better
of the two static splits with one-token pages (49.7\% and 35.0\% against 5.5\% and 6.1\% for Static-2P2D on
Azure (staged) and (mixed)). Both dynamic systems move to 3P1D once, \name within 16\,s of the first
request. \name exceeds Static-3P1D by 8.6 points on Azure (staged) and by 0.6 points on Azure (mixed), and
\cluster-only by 7.0 and 2.0 points. \name's TPOT attainment is 94.6\% and 97.7\%; Static-3P1D and
\cluster-only, which run no stolen work, reach 100\%.

\begin{table}[t]
\centering
\caption{SLO attainment (\%) of Qwen3-32B with TP=2 and 64-token pages at 8\,RPS. \name and \cluster-only
 start from 2P2D.}
\label{tab:app-q32}
\vskip 0.1in
\footnotesize
\setlength{\tabcolsep}{3pt}
\begin{tabular}{@{}l r r r r r r@{}}
\toprule
 & \multicolumn{3}{c}{Azure (staged)} & \multicolumn{3}{c}{Azure (mixed)} \\
\cmidrule(lr){2-4}\cmidrule(l){5-7}
System & All & TTFT & TPOT & All & TTFT & TPOT \\
\midrule
\name & 82.9 & 84.9 & 94.6 & 91.0 & 92.5 & 97.7 \\
\cluster-only & 75.9 & 75.9 & 100.0 & 89.0 & 89.0 & 100.0 \\
Static-3P1D & 74.3 & 74.3 & 100.0 & 90.4 & 90.5 & 100.0 \\
\bottomrule
\end{tabular}
\end{table}

\noindent\textbf{Qwen3-32B load sweep.} On the second server, \name and the static splits run Azure
(staged) at 6, 8 and 10\,RPS (\tref{tab:app-q32-sweep}). \name moves from 2P2D to 3P1D within 54\,s of the
first request and stays there, within 0.8 points of Static-3P1D at 6 and 10\,RPS and 4.3 points above it at 8\,RPS, and above Static-2P2D, the
split it starts from, at every load.

\begin{table}[t]
\centering
\caption{Overall SLO attainment (\%) of Qwen3-32B with TP=2 and 64-token pages on Azure (staged),
second server. \name starts from 2P2D.}
\label{tab:app-q32-sweep}
\vskip 0.1in
\footnotesize
\begin{tabular}{@{}l r r r@{}}
\toprule
System & 6\,RPS & 8\,RPS & 10\,RPS \\
\midrule
\name & 94.3 & 80.1 & 28.4 \\
Static-3P1D & 95.0 & 75.9 & 28.7 \\
Static-2P2D & 49.9 & 9.4 & 0.8 \\
\bottomrule
\end{tabular}
\end{table}

\section{Production Replay}
\label{app:prodreplay}

\noindent We replay 30 minutes of the Azure LLM inference trace~\cite{stojkovic2025dynamollm},
09:04--09:34\,UTC on May 13, 2024, which contains the window of Azure (mixed), on Llama-3.1-8B with the
original timestamps. We thin it per second to $0.9\times$ and $1.1\times$ the highest Azure (mixed) load
that the best static split sustains at 90\% attainment (45\,RPS, 6P2D; \tref{tab:main}). At $0.9\times$
(40.5\,RPS), \name reaches 94.5\% attainment against 95.7\% for 6P2D, 92.8\% for \cluster-only, 37.9\%
for \instance-only, and 0.2\% for Static-4P4D. It moves from 4P4D to 5P3D 8\,s after the first request
and to 6P2D 95\,s later, and holds that split. At $1.1\times$ (49.5\,RPS), 6P2D reaches 26.0\% and \name
13.5\%. \name moves to 5P3D 9\,s after the first request and to 6P2D 205\,s later; at this load even 6P2D
has no spare capacity, so the backlog of those first 214\,s lasts until about 900\,s, after which \name
matches 6P2D. Both runs use \cluster without the projected demand (\appref{app:a2}), which would have moved
it to 6P2D 85\,s ($0.9\times$) and 195\,s ($1.1\times$) earlier.

\section{Role-Switch Correctness and Cost}
\label{app:correctness}

\noindent We exercise \cluster's switch path on Llama-3.1-8B at 4P4D with 64-token pages and \instance
enabled; the controller is off, and a test driver issues the switches. \emph{Idle races} send opposite
switches to two idle workers at once, one race per worker: all 4 races, 8 switches, were accepted.
\emph{Loaded races} send 10 such races and 10 single switches under Azure (mixed) at 35\,RPS: 23 of the
30 switches were accepted, and the 7 refusals were P$\rightarrow$D requests to workers that still ran 1
to 4 residual decode requests (\sref{subsubsec:fluidrole-role-state}). \emph{Residual} switches a decode
worker to prefill under the same load and requests the reverse switch 3\,s later, while residual
requests still run, and again 80\,s later, three times: the worker refused each early request, with 3, 8
and 10 residual requests running, and accepted each later one. \emph{Parity} sends 64 prompts with
greedy decoding and up to 32 output tokens before and after a D$\rightarrow$P and a P$\rightarrow$D
switch: all 64 outputs are identical. In every run, each switch was accepted or refused as planned,
every request completed, every worker's KV accounting and request slots returned to zero, no worker
logged an error while serving, and the workers ended in their launch roles.

\noindent\textbf{Switch cost.}\phantomsection\label{app:switchcost} \tref{tab:switch} breaks down the cost of a role
switch (\sref{subsec:control-switching-overheads}). For the switches issued by \cluster with TP=1, the role
conversion itself takes 3\,ms (D$\rightarrow$P) and 7\,ms (P$\rightarrow$D) at the median and at most 9\,ms;
the rest of a switch is waiting. A D$\rightarrow$P switch waits for the KV transfers and stolen suffixes
already on their way to the worker to land, 289\,ms at the median and at most 1.52\,s; a P$\rightarrow$D
switch waits for its current prefill batch and its outgoing KV transfers, 102\,ms at the median and at most
0.98\,s.

\begin{table}[t]
\centering
\caption{Role-switch cost with 64-token pages: time from the switch request until the worker serves in its new role. Workers use TP=1 except in the TP=2 row. Forced switches are issued by a test driver on Llama-3.1-8B at 4P4D, idle or under Azure (mixed) at 35\,RPS; switches issued by \cluster come from every run of \name and \cluster-only on the first server: Llama-3.1-8B, Qwen3-14B and Qwen3-30B-A3B at TP=1, and Qwen3-32B in the TP=2 row. Rows not marked idle are under load.}
\label{tab:switch}
\vskip 0.1in
\footnotesize
\setlength{\tabcolsep}{4pt}
\begin{tabular}{@{}l >{\raggedright\arraybackslash}p{0.36\columnwidth} l r@{}}
\toprule
Switch & Setting & Time (s) & $n$ \\
\midrule
either & idle, forced & 0.004--0.009 & 8 \\
P$\rightarrow$D & issued by \cluster & 0.11 median, $\le$0.99 & 28 \\
P$\rightarrow$D & forced & 0.008--1.04 & 8 \\
P$\rightarrow$D & forced, with stealing & 0.008--0.58 & 8 \\
P$\rightarrow$D & idle, forced, graphs captured at the switch & 3.65--3.69 & 4 \\
D$\rightarrow$P & issued by \cluster & 0.29 median, $\le$1.52 & 153 \\
D$\rightarrow$P & forced & 0.015--0.90 & 8 \\
D$\rightarrow$P & forced, with stealing & 0.015--0.42 & 8 \\
D$\rightarrow$P & forced, KV pool capped at 1/3 & 0.016--0.55 & 6 \\
D$\rightarrow$P & forced, draining (no residual requests) & 2.2--11.5 & 8 \\
D$\rightarrow$P & issued by \cluster, TP=2 (Qwen3-32B) & 0.029--1.31 & 4 \\
residual & requests left after a D$\rightarrow$P issued by \cluster & 16.7 median, $\le$62.2 & 149 \\
\bottomrule
\end{tabular}
\end{table}

\section{Additional Experiments}
\label{app:extra}

\subsection{Main Comparison Details}
\label{app:extra-main}

\noindent \tref{tab:main} lists the overall SLO attainment of every cell of the main comparison
(\fref{fig:eval-main-staged-azure-rps-sweep}) except Qwen3-30B-A3B (\tref{tab:app-q30}), together
with the other static splits run at each load.

\noindent\textbf{Second server.} These runs use a second server of the same type (8$\times$A100-SXM4-80GB
with NVLink): every ShareGPT (staged) run; every Qwen3-14B \name run with the 150\,ms TPOT SLO and the static runs at
22\,RPS; the Qwen3-30B-A3B runs with measured decode costs and the Qwen3-32B load sweep (\appref{app:models}); the six-GPU runs (\appref{app:thirdparty}); and the Llama-3.1-8B
runs on Azure (staged) at 50 and 55\,RPS and from 6P2D. The static Qwen3-14B runs on Azure (staged) at
15, 20 and 25\,RPS are first-server runs scored at 3000/150\,ms from their per-request latencies, since
a static split serves requests the same way under any SLO; their attainment changes by at most
0.01 points from 3000/100\,ms. Three configurations run on both servers reach 1.1--2.0 points less on
the second: on Llama-3.1-8B on Azure (staged) at 40\,RPS, \name reaches 92.5\% against 93.6\% and
Static-6P2D 69.4\% against 70.6\%; on Qwen3-14B at 20\,RPS, Static-6P2D reaches 81.9\% against 83.9\%.

\noindent\textbf{ShareGPT (staged).} The workload keeps the three stages and the Azure-code stage of
Azure (staged) and takes the lengths of its two conversation stages from ShareGPT
conversations~\cite{sharegpt}, in the order SGLang's serving benchmark samples them. A
request is the last user turn of a conversation: its prompt is the conversation up to and including that
turn, and its output is the reply. Each conversation thus contributes one request with its full history,
as a long chat session does at its latest turn; the prompts average 1,085 tokens (median 1,267) and the
outputs 337. ShareGPT's released conversations are split into pieces of about 2K tokens, so 98\% of
the prompts stay at or below 2,048 tokens. The 6K-request stages of Qwen3-14B take the first 12K of the same conversations (1,087 prompt and 335
output tokens on average). Lengths are counted with the Llama-3.1 tokenizer and filtered as in SGLang's serving
benchmark, including its 8,192-token context limit.

\noindent\textbf{Starting split.} For Llama-3.1-8B, starting from 5P3D instead of 4P4D, \name reaches
99.2\% and 98.1\% on Azure (staged) at 30 and 35\,RPS, against 99.1\% and 97.8\% from 4P4D. Starting
from 6P2D at 45\,RPS, it reaches 74.9\% on the second server, against 73.7\% from 4P4D.

\noindent\textbf{Beyond 45\,RPS.} On Llama-3.1-8B on Azure (staged) at 50 and 55\,RPS (second server),
every configuration stays below 33\%: \name reaches 29.3\% and 21.4\%, Static-6P2D 32.0\% and 28.3\%,
and Static-4P4D 0.9\% at 50\,RPS. Both pools are short at these loads: at 50\,RPS, the prefill demand of
the last stage exceeds what six prefill workers serve at the target utilization, so \cluster keeps 7P1D
for 225\,s of that stage while requests queue for the single decode worker, the overload of
\tref{tab:app-tight}.

\noindent\textbf{Llumnix.}\phantomsection\label{app:llumnix} We run Llumnix at commit aa9097c (with
llumnix-kv 5f8dd15) on vLLM~0.18.0 in its official adaptive-P/D deployment~\cite{llumnix-repo}: SLO-aware
scheduling with adaptive P/D and full-mode scheduling, dispatch thresholds of 0.9, rescheduling every
100\,ms, and request migration. Each of the eight A100-80GB GPUs of an NVLink server with eight
200\,Gb/s InfiniBand NICs runs one instance (TP=1), starting as 4P4D. As in the official deployment, the
engines run in eager mode, without CUDA graphs, with a GPU-memory utilization of 0.85, and KV caches move
between instances over RDMA through Llumnix's KVT connector. The scheduler's TTFT target is the TTFT SLO,
and its latency profiles were measured on the same server. Its TPOT target is 1000\,ms instead of the
100\,ms SLO. Its dispatcher places a request only where the predicted latency stays below 0.9 times the
target; with the SLO as the target, it rejects every prompt longer than about 2K tokens (34\% of the
Llama-3.1-8B requests of Azure (mixed)), most Llama-3.1-8B requests receive no first token within 60\,s,
and only 2.3\% of them meet both SLOs on Azure (staged) at 25\,RPS, and 12.2\% and 1.4\% on Azure (mixed)
at 35 and 55\,RPS. Of the targets we tried, 100, 200, 400 and 1000\,ms, 1000\,ms is the first that admits
the whole prompt-length distribution. Under adaptive P/D, with migration enabled, its role logs show it at
about 6P2D within the first minute of each run and at that split for the rest of the run. An older copy of our benchmark client sends the same requests with the same
Poisson arrival process and abandons a request after 60\,s without a first token, or without a new token
once it streams; we score the runs as ours (\sref{subsec:eval-method}). Llumnix loses attainment on TTFT:
over completed requests, its TPOT attainment is at least 97\% in every cell, while at 40\,RPS on Azure
(staged), 9,422 of the 30,000 Llama-3.1-8B requests receive no first token within 60\,s.

\begin{table*}[t]
\centering
\caption{Main comparison with 64-token pages (overall SLO attainment, \%, seed 1). The
best static split is the best of the static splits run at each load; it is 6P2D unless another split is
given. Qwen3-14B is scored at 3000/150\,ms on the staged workloads and at 3000/100\,ms on Azure (mixed).
Llumnix runs its adaptive-P/D deployment from 4P4D on the Azure workloads. Other splits run at a load are
listed with their attainment; -- means not run. Workloads: staged = Azure (staged), ShareGPT = ShareGPT (staged), mixed = Azure (mixed).}
\label{tab:main}
\vskip 0.1in
\footnotesize
\begin{tabular}{@{}l r r r r r l r@{}}
\toprule
Model, workload & RPS & \name & Best static & Static-4P4D & Llumnix & Other static splits & Oracle \\
\midrule
Llama-3.1-8B, staged & 25 & 99.7 & 99.4 & 75.7 & 100.0 & -- & -- \\
 & 30 & 99.1 & 98.2 & 39.5 & 91.1 & 5P3D 87.2 & -- \\
 & 35 & 97.8 & 92.6 & 27.4 & 82.6 & 5P3D 57.5 & 98.2 \\
 & 40 & 93.6 & 70.6 & 2.4 & 38.9 & -- & 95.3 \\
 & 45 & 73.7 & 40.5 & 1.2 & 30.7 & -- & 85.2 \\
\midrule
Llama-3.1-8B, ShareGPT & 30 & 98.5 & 97.9 & 56.4 & -- & 5P3D 86.5 & -- \\
 & 35 & 96.9 & 63.0 (5P3D) & 51.3 & -- & 6P2D 19.3 & -- \\
 & 40 & 92.8 & 45.0 (4P4D) & 45.0 & -- & 5P3D 34.1, 6P2D 5.5 & -- \\
 & 45 & 64.5 & 33.2 (4P4D) & 33.2 & -- & 5P3D 28.0, 6P2D 4.0 & -- \\
\midrule
Qwen3-14B, staged & 15 & 98.5 & 98.3 & 56.2 & 80.8 & 5P3D 93.1 & -- \\
 & 20 & 95.3 & 83.9 & 20.5 & 30.6 & 5P3D 43.5 & -- \\
 & 22 & 90.1 & 59.6 & 2.2 & -- & -- & -- \\
 & 25 & 53.8 & 33.4 & 1.3 & 4.4 & 5P3D 24.2 & -- \\
\midrule
Qwen3-14B, ShareGPT & 16 & -- & 97.1 & -- & -- & 5P3D 86.4 & -- \\
 & 18 & 96.5 & 78.8 & 55.4 & -- & 5P3D 66.0 & -- \\
 & 20 & 92.8 & 52.2 (5P3D) & 51.7 & -- & 6P2D 8.1 & -- \\
 & 22 & 85.1 & 46.6 (4P4D) & 46.6 & -- & 5P3D 39.8, 6P2D 5.5 & -- \\
 & 24 & 58.4 & 39.6 (4P4D) & 39.6 & -- & 5P3D 33.9, 6P2D 4.4 & -- \\
\midrule
Llama-3.1-8B, mixed & 35 & 98.6 & 99.0 & 19.1 & 87.0 & 5P3D 96.1 & -- \\
 & 40 & 95.8 & 97.7 & 6.1 & 84.7 & -- & -- \\
 & 45 & 89.2 & 94.1 & 3.3 & 51.0 & 5P3D 19.1 & -- \\
 & 50 & 58.8 & 73.9 & 1.6 & 20.3 & 7P1D 7.3 & -- \\
 & 55 & 4.2 & 19.0 & 1.6 & 6.1 & 7P1D 6.9 & -- \\
\midrule
Qwen3-14B, mixed & 15 & 99.4 & 99.9 & 93.8 & 99.8 & 5P3D 99.0 & -- \\
 & 20 & 97.0 & 97.9 & 24.2 & 28.4 & 5P3D 91.4 & -- \\
 & 25 & 85.0 & 88.3 & 5.0 & 28.0 & 5P3D 29.4 & -- \\
\bottomrule
\end{tabular}
\end{table*}

\subsection{Component Ablation on Azure (mixed)}
\label{app:mixed}

\begin{figure}[t]
\centering
\includegraphics[width=\columnwidth]{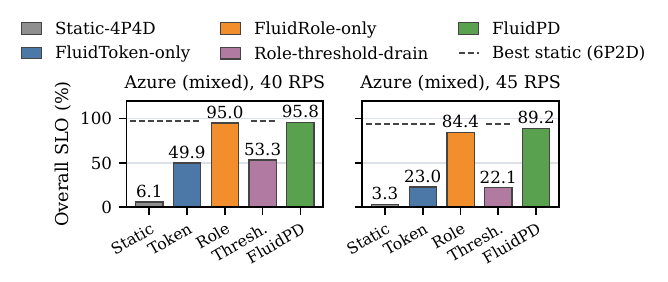}
\caption{Component ablation on Llama-3.1-8B on Azure (mixed), starting from 4P4D. Bars and line as in
\fref{fig:analysis-ablation}.}
\label{fig:app-ablation-mixed}
\end{figure}

\noindent The component ablation on Azure (mixed) (\fref{fig:app-ablation-mixed}) shows the same order as on
Azure (staged). \name is 4.8 points above \cluster-only at 45\,RPS, and 4.5 of these points come from
requests that arrive in the first 40\,s, while the allocation is still adapting.

\subsection{Seed Repeats}

\noindent Each configuration runs once with arrival seed 1 (\sref{subsec:eval-method}). \tref{tab:seeds}
repeats seven cells near the knees of the main comparison with seed 2. \name's attainment differs
between the seeds by at most 3.4 points, except for Qwen3-14B on Azure (staged) at 25\,RPS, where the load
sits on the steep part of the curve and the seeds give 53.8\% and 46.2\%; Static-6P2D's differs by at
most 2.2 points, and \name's difference to the best static split has the same sign for both seeds in
every cell.

\begin{table*}[t]
\centering
\caption{Overall SLO attainment (\%) with arrival seeds 1 and 2 (64-token pages), and $\Delta$, \name minus
Static-6P2D, the best static split in every cell, in points from unrounded values. Qwen3-14B at 3000/150\,ms, with \name on the
second server (\appref{app:extra-main}).}
\label{tab:seeds}
\vskip 0.1in
\footnotesize
\begin{tabular}{@{}l r r r r r r r@{}}
\toprule
 & & \multicolumn{2}{c}{\name} & \multicolumn{2}{c}{6P2D} & \multicolumn{2}{c}{$\Delta$} \\
\cmidrule(lr){3-4}\cmidrule(lr){5-6}\cmidrule(l){7-8}
Model, workload & RPS & s1 & s2 & s1 & s2 & s1 & s2 \\
\midrule
Llama-3.1-8B, staged & 35 & 97.8 & 97.9 & 92.6 & 92.2 & $+5.2$ & $+5.7$ \\
 & 40 & 93.6 & 92.5 & 70.6 & 68.9 & $+23.0$ & $+23.6$ \\
 & 45 & 73.7 & 70.3 & 40.5 & 41.6 & $+33.2$ & $+28.7$ \\
Llama-3.1-8B, mixed & 40 & 95.8 & 95.7 & 97.7 & 97.9 & $-1.9$ & $-2.2$ \\
 & 45 & 89.2 & 89.4 & 94.1 & 93.7 & $-4.9$ & $-4.3$ \\
Qwen3-14B, staged & 20 & 95.3 & 94.9 & 83.9 & 81.7 & $+11.3$ & $+13.2$ \\
 & 25 & 53.8 & 46.2 & 33.4 & 33.1 & $+20.4$ & $+13.1$ \\
\bottomrule
\end{tabular}
\end{table*}

\subsection{TTFT and TPOT Attainment}
\label{app:ttft-tpot}

\begin{figure*}[t]
    \centering
    \begin{subfigure}[t]{0.49\textwidth}
        \centering
        \includegraphics[width=\linewidth]{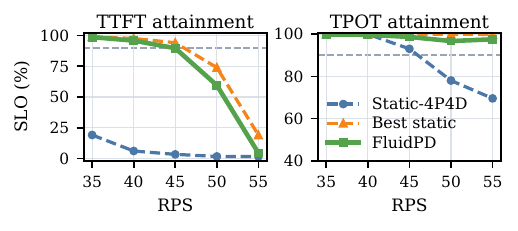}
        \caption{TTFT and TPOT contribution on Llama-3.1-8B for Azure (mixed).}
        \label{fig:eval-ttft-tpot-contribution}
    \end{subfigure}
    \hfill
    \begin{subfigure}[t]{0.49\textwidth}
        \centering
        \includegraphics[width=\linewidth]{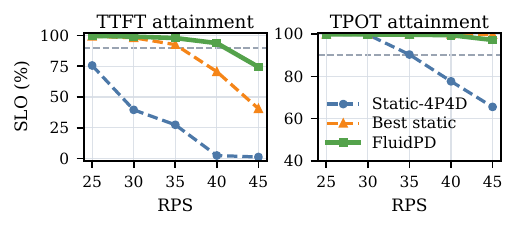}
        \caption{TTFT and TPOT contribution on Llama-3.1-8B for Azure (staged).}
        \label{fig:eval-ttft-tpot-contribution-staged}
    \end{subfigure}
    \caption{TTFT and TPOT attainment on Llama-3.1-8B.}
    \label{fig:eval-overview}
\end{figure*}

\noindent \fref{fig:eval-overview} breaks down the overall SLO attainment of Llama-3.1-8B on Azure (mixed)
and Azure (staged) into TTFT and TPOT attainment. Overall attainment follows TTFT attainment for every
system (\sref{subsec:eval-main}). Static-4P4D's TPOT attainment falls with the
load, to 65.6\% at 45\,RPS on Azure (staged), because requests abandoned at the 60\,s first-token limit
count as violations; nearly all of its completed requests meet the TPOT SLO.

\subsection{Sensitivity to SLO Targets}
\label{app:slo}\label{subsec:slo-sensitivity}

\begin{figure}[!t]
    \centering
    \includegraphics[width=\linewidth]{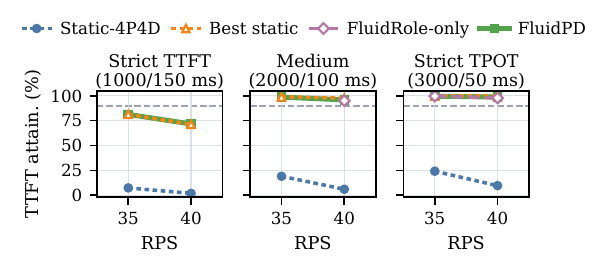}
    \caption{Sensitivity to SLO targets on Azure (mixed) for Llama-3.1-8B (TTFT/TPOT SLO in ms). We omit TPOT because it remains close to 100\%. Static points at 1000/150 and 3000/50 rescore the per-request latencies of the 2000/100 runs, since a static split serves requests the same way under any SLO; \cluster-only is shown at 3000/50 and, at 40\,RPS, at 2000/100.}
    \label{fig:analysis-slo-targets}
\end{figure}

\noindent We evaluate Llama-3.1-8B on Azure (mixed) at 35 and 40\,RPS for three latency profiles (\fref{fig:analysis-slo-targets}): strict TTFT (1000\,ms TTFT / 150\,ms TPOT), medium (2000\,ms TTFT / 100\,ms TPOT), and strict TPOT (3000\,ms TTFT / 50\,ms TPOT); \name's workers and controller run with the SLOs of each profile. We focus on TTFT because TPOT attainment remains at or above 98.8\% at these operating points. Starting from 4P4D, \name moves to 6P2D under every profile and reaches the TTFT attainment of the best static split, which is 6P2D at every point: within 0.3 points at 35\,RPS, and between 1.8 points below and 0.5 points above at 40\,RPS, while Static-4P4D stays below 25\%. Under the strict 50\,ms TPOT, the TPOT slack that \instance may use shrinks; \name matches \cluster-only at 35\,RPS and exceeds it by 1.0 points at 40\,RPS.

\subsection{Pulses of Different Lengths}

\noindent A base stream of Azure (mixed) requests at 18\,RPS, $0.6\times$ the highest load at which
Static-4P4D sustains 90\% attainment with 64-token pages (30\,RPS), runs for $120 + d + 180$\,s; during
a pulse of $d$ seconds, Azure-code requests arrive at another 18\,RPS. Requests arrive at uniformly random
times within each phase. \tref{tab:app-pulses} shows Llama-3.1-8B starting from 4P4D. A 5\,s pulse is
absorbed by either mechanism alone. As the pulse lengthens, the two mechanisms diverge: at $d=20$\,s
\instance-only reaches 94.5\% and \cluster-only 90.6\%, while at $d=300$\,s \instance-only reaches
53.5\% and \cluster-only 93.5\%. \name has the highest attainment at every pulse length, from 92.4\% to
99.0\%.

\begin{table}[t]
\centering
\caption{Overall SLO attainment (\%) under pulses of $d$ seconds on Llama-3.1-8B with 64-token pages,
starting from 4P4D.}
\label{tab:app-pulses}
\vskip 0.1in
\footnotesize
\setlength{\tabcolsep}{4pt}
\begin{tabular}{@{}r r r r r@{}}
\toprule
$d$ (s) & \makecell[r]{\instance-\\only} & \makecell[r]{\cluster-\\only} & \name & Static-4P4D \\
\midrule
5 & 98.6 & 98.9 & 99.0 & -- \\
20 & 94.5 & 90.6 & 95.1 & -- \\
40 & 88.1 & 90.1 & 94.1 & -- \\
90 & 76.0 & 86.4 & 92.4 & -- \\
300 & 53.5 & 93.5 & 96.2 & 20.4 \\
\bottomrule
\end{tabular}
\end{table}

\subsection{Both Pools Under Pressure}

\noindent Two synthetic workloads on Llama-3.1-8B mix a prompt-heavy stream (4,000 prompt and 16 output
tokens per request) and a decode-heavy stream (512 and 1,000), with requests at uniformly random times
within each phase. Their rates are set relative to the throughput of four prefill workers (58.9K prompt
tokens/s) and of four decode workers (3.2K output tokens/s) with 64-token pages. The \emph{step}
workload runs both streams at $0.5\times$ for 60\,s, $1.4\times$ for 90\,s, and $0.5\times$ for 60\,s,
so that both pools are overloaded at once. The \emph{burst} workload runs the decode-heavy stream at
$0.4\times$ for 180\,s and adds the prompt-heavy stream at $1.5\times$ for 20\,s from 60\,s on. Besides
\instance-only and \cluster-only, \emph{Independent} combines \instance without its decode-load limit
($\mathrm{DPI}^{\mathrm{high}}_{\mathrm{load}}$, \tref{tab:params}) with the threshold controller of
Role-threshold-drain (\sref{subsec:eval-ablation}), here with non-draining switches. All systems start from 4P4D (\tref{tab:app-tight}).
On the step, \name reaches 49.8\%, 8.7 points above Independent and 14.3 points above \instance-only.
On the burst, \name and \instance-only both reach 39.7\%, 0.9 points above Independent.

\begin{table}[t]
\centering
\caption{Overall SLO attainment (\%) on the step and burst workloads (Llama-3.1-8B, 64-token pages, all
systems starting from 4P4D).}
\label{tab:app-tight}
\vskip 0.1in
\footnotesize
\begin{tabular}{@{}l r r@{}}
\toprule
System & Step & Burst \\
\midrule
Static-4P4D & 18.0 & 33.9 \\
\instance-only & 35.5 & 39.7 \\
\cluster-only & 28.6 & 37.2 \\
Independent & 41.1 & 38.8 \\
\name & 49.8 & 39.7 \\
\bottomrule
\end{tabular}
\end{table}

\section{Comparison with vLLM and Dynamo}
\label{app:thirdparty}

\begin{figure}[t]
\centering
\makebox[\linewidth][c]{\includegraphics[width=\linewidth]{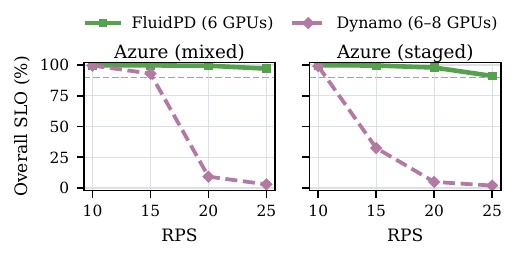}}
\caption{\name on a fixed 6-GPU budget versus Dynamo autoscaling, which starts from the same 6-GPU
3P3D deployment and may grow to 8 GPUs, on Llama-3.1-8B with one-token KV pages.}
\label{fig:app-thirdparty-autoscaling}\label{fig:analysis-autoscaling}%
\end{figure}

\begin{figure}[t]
\centering
\includegraphics[width=\columnwidth]{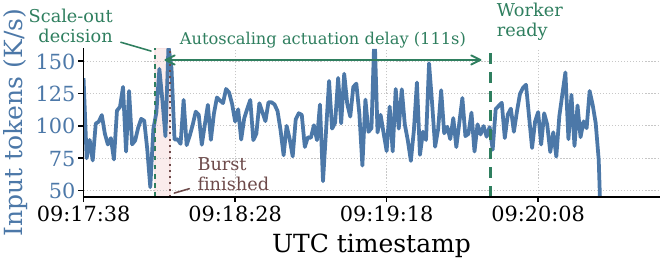}
\caption{A new Dynamo worker becomes ready a median of 111\,s after the scale-out decision, long after the burst ends. The curve is the input-token demand of Azure (mixed) at its trace timestamps; the decision is placed at the start of the burst.}
\label{fig:autoscaling-short-term-downside}
\end{figure}

\noindent This appendix gives the setup of the vLLM and Dynamo runs in \fref{fig:eval-main-staged-azure-rps-sweep}
and compares \name with Dynamo's autoscaling. vLLM~0.25.1 runs with its NIXL connector, 128-token KV blocks, eager execution and a GPU memory utilization of 0.7. Dynamo~v1.3.0
runs its SGLang backend with one-token KV pages on Kubernetes, one pod per worker, with a static memory
fraction of 0.75 and KV transfers over NIXL's libfabric backend on EFA. Both route requests round-robin and
replay the trace files and request counts of \sref{subsec:eval-method}. Their benchmark client has no 60\,s
first-token limit, so no request is abandoned, and it does not force output lengths; the
generated lengths still equal the traces'. We rescore every request with the scoring of our client.

\noindent\textbf{Autoscaling.} \fref{fig:app-thirdparty-autoscaling} compares \name with Dynamo's
autoscaling on Llama-3.1-8B, to determine whether adding workers addresses the same P/D imbalance that
\name resolves within the running deployment. Both start from the same 6-GPU 3P3D deployment; \name
remains within these six GPUs, whereas Dynamo may scale the deployment to eight. We run Dynamo's planner
with \texttt{optimization\_target: latency} and leave its remaining parameters at their defaults; its
\texttt{sla} target would match our SLO-based setting more closely, but it was not stable in the version
we evaluated. Dynamo's autoscaling runs use the client of its static runs. \name runs with
one-token KV pages, as Dynamo's SGLang backend does, on the second server (\appref{app:extra-main}); with
one-token pages, \cluster keeps at least two of the six workers in the decode role
(\appref{app:pagesize}). On Azure (mixed), the dominant event is a short prefill-heavy burst. At 20\,RPS,
\name achieves 99.5\% overall SLO attainment on six GPUs, compared with 9.2\% for Dynamo. A worker that
Dynamo adds becomes ready a median of 111\,s after the planner decides to scale out
(\sref{subsec:autoscaling-limitations}, \fref{fig:autoscaling-short-term-downside}), longer than the
burst and the TTFT violations it causes. Kubernetes creates the pod within about 3\,s of the decision;
the rest of the delay is runtime initialization, weight loading and CUDA-graph capture. \name instead
responds within the existing deployment: \cluster switches one decode worker to prefill (3P3D to 4P2D)
within 42\,s of the first request, and \instance runs bounded prefill work on decode-side slack without
waiting for additional workers. Azure (staged) contains a longer-lasting demand shift and is therefore
more favorable to autoscaling. At 20\,RPS, \name achieves 98.0\% overall SLO attainment, whereas Dynamo
achieves 4.9\%; at 25\,RPS, \name still reaches 97.1\% on Azure (mixed) and 91.2\% on Azure (staged).
Under Dynamo's planner, the deployment repeatedly changes its P/D allocation, including removing workers
and later adding them back, and does not converge quickly to a configuration that matches the workload.

\section{KV Page Size}
\label{app:pagesize}
\label{subsec:eval-pagesize}

\noindent This appendix repeats the Llama-3.1-8B configurations of the main comparison on the Azure workloads with one-token KV pages and reports one-token results for Qwen3-32B and Qwen3-30B-A3B. In the one-token-page
runs, \cluster keeps at least $\lceil N/4 \rceil$ of the $N$ workers in the decode role and switches by
$\delta_t$ alone (\sref{subsec:fluidrole_policy}).

In P/D disaggregation, the KV of every prefilled request moves to a decode worker, and the cost of that move depends on how the decode worker's KV slots are laid out. In SGLang v0.5.4 with NIXL~0.6.0, on which \name is built, the prefill scheduler posts each KV transfer itself, right after the prefill batch that produced the KV, and NIXL holds Python's global interpreter lock while it posts. A transfer carries one descriptor for every run of KV slots that is contiguous on both the prefill and the decode worker, for every layer and separately for K and V, so a request whose KV lands in $r$ such runs costs $2\ell r$ descriptors on a model with $\ell$ layers. With one-token pages, a decode worker allocates its KV slot by slot: every decode step takes one slot for each running request, and freed slots return to the free list unsorted until it runs short. As the decode pool fills, the slots of a newly admitted request scatter across the pool and $r$ grows toward the prompt length. Posting then costs time per prompt token on the thread that schedules prefill, and this cost grows with the decode worker's KV usage. With pages of $P$ tokens, every page is contiguous on both sides, so a prompt of $L$ tokens takes at most $\lceil L/P \rceil$ runs whatever the state of the free list.

\begin{figure}[t]
    \centering
    \includegraphics[width=\linewidth]{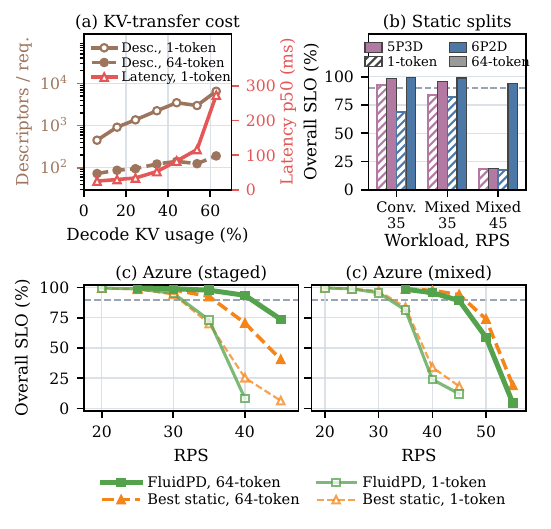}
    \caption{KV page size on Llama-3.1-8B. (a) Static deployments with one-token pages at 35\,RPS: NIXL descriptors per request, from a replay of the decode workers' KV allocations, and median KV-transfer latency versus the decode worker's KV usage; the dashed line replays the same allocations with 64-token pages. (b) Static 5P3D and 6P2D on the conversation stage at 35\,RPS and on Azure (mixed) at 35 and 45\,RPS, with one-token (hatched) and 64-token (solid) pages. (c) \name versus the best static split at each page size; hollow markers show one-token pages.}
    \label{fig:eval-pagesize}
\end{figure}

\fref{fig:eval-pagesize}(a) shows the effect on Llama-3.1-8B static deployments with one-token pages. The descriptors posted per request rise from 1,244 at 19\% mean decode-pool usage (conversation stage, 5P3D) to 3,615 at 41\% (6P2D), the KV-transfer latency follows the descriptor count (Spearman $\rho=0.80$), and each prefill scheduler stops for 3.2\,s and 15.0\,s per minute in stalls of 1\,s or more, mostly while it posts. With 64-token pages, the same allocations take 88 and 139 descriptors per request.

The cost falls on decode-light splits, whose fewer decode workers carry the decode load in fuller pools. The page size therefore changes which split serves a workload best, which is the quantity \cluster adapts. \fref{fig:eval-pagesize}(b) compares the static splits at both page sizes: on Azure (mixed), the best static split of Llama-3.1-8B is 5P3D with one-token pages and 6P2D with 64-token pages, and on the conversation stage alone at 35\,RPS, 6P2D rises from 69.0\% to 99.7\% while 5P3D moves from 92.5\% to 98.6\%. \fref{fig:eval-pagesize}(c) compares \name with the best static split at each page size, with \cluster configured for each page size as described above. With one-token pages, \name and the best static split both sustain 90\% attainment up to 30\,RPS on Azure (staged) and (mixed). With 64-token pages, \name sustains 40\,RPS on both workloads, against 35\,RPS on Azure (staged) and 45\,RPS on Azure (mixed) for the best static split.

The same holds beyond Llama-3.1-8B. With one-token pages, Qwen3-32B at TP=2 reaches 8.9\% and 24.9\% with \name and 49.7\% and 35.0\% with Static-3P1D on Azure (staged) and (mixed) at 8\,RPS; with 64-token pages, \name reaches 82.9\% and 91.0\% against 74.3\% and 90.4\%. For Qwen3-30B-A3B at 20\,RPS, moving from one-token to 64-token pages raises \name from 40.0\% to 96.6\% on Azure (mixed) and from 6.7\% to 93.6\% on Azure (staged).

SGLang's default page size on A100 GPUs is still one token in v0.5.22, while its P/D deployment examples use 64-token pages~\cite{sglang-pd-lws}. Since v0.5.12, SGLang posts NIXL transfers from background threads, which takes posting off the prefill scheduler. The allocator and the descriptor count are unchanged, so with one-token pages the number of descriptors per transfer still grows with decode-pool usage.

\end{document}